# Hybrid Bayesian network discovery with latent variables by scoring multiple interventions

Kiattikun Chobtham[1], Anthony C. Constantinou[1] and Neville K. Kitson[1].

1. Bayesian Artificial Intelligence research lab, Risk and Information Management (RIM) research group, School of Electronic Engineering and Computer Science, Queen Mary University of London (QMUL), London, UK, E1 4NS.
   E-mails: k.chobtham@qmul.ac.uk, a.constantinou@qmul.ac.uk, and n.k.kitson@qmul.ac.uk.

**Abstract**

In Bayesian Networks (BNs), the direction of edges is crucial for causal reasoning and inference. However, Markov equivalence class considerations mean it is not always possible to establish edge orientations, which is why many BN structure learning algorithms cannot orientate all edges from purely observational data. Moreover, latent confounders can lead to false positive edges. Relatively few methods have been proposed to address these issues. In this work, we present the hybrid mFGS-BS (majority rule and Fast Greedy equivalence Search with Bayesian Scoring) algorithm for structure learning from discrete data that involves an observational data set and one or more interventional data sets. The algorithm assumes causal insufficiency in the presence of latent variables and produces a Partial Ancestral Graph (PAG). Structure learning relies on a hybrid approach and a novel Bayesian scoring paradigm that calculates the posterior probability of each directed edge being added to the learnt graph. Experimental results based on well-known networks of up to 109 variables and 10k sample size show that mFGS-BS improves structure learning accuracy relative to the state-of-the-art and it is computationally efficient.

## 1. Introduction

A Bayesian Network (BN) is a probabilistic graphical model with a Directed Acyclic Graph (DAG) $G$ where nodes $\mathbf{X} = \{X_1, \dots, X_N\}$ represent random variables and directed edges represent dependencies or causal relationships between variables (Verma and Pearl, 1990). A BN is a generative model that captures the joint probability distribution of the data variables. The dependencies between discrete variables are described via conditional probabilities, such as $P(X_i|\text{parent}(X_i))$ where $\text{parent}(X_i)$ is the set of parents of node $X_i$ in the DAG. The joint distribution over all nodes is defined as the product of all conditional probabilities as follows:

$$P(X_1, X_2, X_3, \dots X_N) = \prod_{i=1}^{N} P(X_i|\text{parent}(X_i))$$

Because directed edges in BNs can often be viewed as causal relationships, BNs offer the potential to go beyond predictive inference by enabling causal reasoning for intervention and counterfactual reasoning. Examples of BNs applied to different areas include medicine (Thornley et al., 2012), sports (Constantinou, 2020), social science (de Waal et al., 2016), finance (Constantinou and Fenton, 2017), geology (Runge et al., 2019), bioinformatics (Sachs et al., 2005) and law (de Zoete et al., 2019). Many of the BNs applied to real-world problems are determined by knowledge, or both knowledge and data (Constantinou et al., 2016). In this paper, however, we focus on the automated discovery of BN structures from data.

Structure learning methods generally fall into two main classes of learning known as score-based and constraint-based learning. The score-based algorithms rely on search methods that explore the search space of graphs and an objective function that scores each graph visited, where the highest scoring graph discovered is returned as the preferred graph. On the other hand, constraint-based learning relies on conditional independence (CI) tests that are used to determine edges and the orientation of some of those edges. Hybrid learning algorithms that combine the two above approaches are often viewed as an additional category of learning. Irrespective of the learning class, BN structure learning represents an NP-hard problem where the number of possible graphs grows super-exponentially with the number of variables. Moreover, large or dense networks tend to require large sample sizes to achieve reasonable structure learning accuracy, and this is a problem because computational complexity increases both with the number of the variables and the sample size.

Learning the structure of a BN involves two more important issues that go beyond computational complexity. Firstly, structures that represent a serial connection (A → B → C or A ← B ← C) or a divergence connection (A ← B → C) cannot be differentiated by observational data, which means algorithms may fail to orientate these edges. This is because these structures encode the same CI statement A ⊥ C | B. Randomly orientating these edges into one of the equivalence structures leads to different DAGs. This set of DAGs is known as a Markov equivalence class and is represented by a Completed Partially DAG (CPDAG). Secondly, data often do not capture all the relevant variables, and learning from data with latent variables is referred to as learning under the assumption of causal insufficiency. A latent confounder represents a special case of a latent variable where the missing variable is a common cause of two or more observed variables, and this tends to lead to spurious edges between observed variables. Because a DAG is not detailed enough to capture spurious relationships, ancestral graphs have been proposed for this purpose. Specifically, the Maximal Ancestral Graph (MAG) by Richardson and Spirtes (2000) represents an extension of the DAG where directed edges represent parental or ancestral relationships and bidirected edges represent confounding. Moreover, a Partial Ancestral Graph (PAG) represents a set of Markov equivalent MAGs (Spirtes et al., 2001), in the same way that a CPDAG represents a set of Markov equivalent DAGs.

Numerous constraint-based algorithms have been proposed to tackle learning under the assumption of causal insufficiency from purely observational data. Well-established constraint-based algorithms include the FCI algorithm by Spirtes et al. (2001), and its variants, conservative FCI (cFCI) by Ramsey et al. (2012), majority rule FCI (mFCI) by Colombo and Maathuis (2014), and RFCI by Colombo et al. (2011). The FCI algorithm assumes that the joint probability distribution is a perfect map with a faithful graph, but this assumption is often violated when applying the algorithm to real data. The cFCI, mFCI and RFCI are all FCI-based variants. For example, the RFCI algorithm can be viewed as a faster version of FCI that performs fewer CI tests. Details about the cFCI and mFCI algorithms are provided in subsection 2.5.

Hybrid algorithms that learn under the assumption of causal insufficiency include CCHM (Chobtham and Constantinou, 2020), M$^3$HC (Tsirlis et al., 2018), RFCI-BSC (Jabbari et al., 2017) and GFCI (Ogarrio et al., 2016). The CCHM algorithm combines the first and second steps of cFCI with a greedy hill-climbing search similar to M$^3$HC, and uses causal effects to return a MAG. Both CCHM and M$^3$HC assume the data follow a Gaussian distribution. The GFCI algorithm works with both discrete and continuous variables. It combines the score-based FGS (Ramsey, 2015) with the orientation rules in FCI. It starts by obtaining the dependencies from the learnt CPDAG returned by FGS, and performs CI tests on those dependencies to remove potential false positive edges. The result of this process is a skeleton. Finally, orientation rules of FCI are applied to the graph skeleton to produce a PAG. RFCI-BSC is the most relevant algorithm to our work and is discussed in subsection 2.4.

Structure learning algorithms that learn purely from observational data are restricted to identifying graphs up to Markov equivalence classes. This means interventional data is often required to identify edge orientations, and this can be achieved by comparing post-interventional distributions with pre-interventional distributions. Classic randomised controlled trials (Fisher, 1935) can be viewed as one kind of interventional data that captures treatments and their outcomes. They typically involve randomly assigning patients into two groups, where the so-called treatment group is given the drug being tested, and the control group is given a placebo. If the outcome distribution differs significantly between the two groups, the difference is viewed as the effect of the drug. Pearl (Pearl, 2000) describes this as the difference between "*given that we see*" (observational data) and "*given that we do*" (interventional data). Therefore, interventional data can be used in conjunction with observational data to orientate edges that would otherwise remain unoriented.

Algorithms that learn from both observational and interventional data tend to do so from pooled data, which is a method that pools all data sets together with intervened variables specified. These algorithms aim to generate a graph that is consistent, as much as possible, with all input data. Examples include IGSP (Wang et al., 2017) and GIES (Hauser and Bühlmann, 2012) that return a DAG from pooled causally sufficient data. Other methods involve determining the results of CI tests from each data set separately and constructing a single graph using conflict resolution strategies. For causally insufficient data, the COmbINE algorithm by Triantafillou and Tsamardinos (2015) implements the cFCI approach to learn the common characteristics and the results of CI tests from different data sets, which it then converts into Boolean Satisfiability (SAT) instances in a MINISAT application to resolve any conflicts. Other algorithms that operate on such results of CI tests include HEJ (Hyttinen et al., 2014) which uses Clingo (Gebser et al., 2011) - an Answer Set Programming (ASP) rule-based declarative programming language that solves various representations of NP-hard optimisation tasks (Gelfond and Lifschitz, 1988; Niemela, 1999) – for conflict resolution. It produces cyclic directed mixed graphs encoding results of CI tests from conditioning and marginalisation operations, and the graphs may contain directed, bidirected or undirected edges. The ACI algorithm (Magliacane et al., 2017) also relies on Clingo and can be viewed as a computationally less expensive variant of HEJ that operates in the search space of ancestral graphs but which does not support bidirected edges for latent confounder representation. Lastly, JCI (Mooij et al., 2020) is a constraint-based algorithm that uses auxiliary context variables and system variables, which the authors define as variables of interest (presumably observed variables) and intervention targets respectively. JCI learns from a pooled data set including knowledge about the relationship between context variables and generates a directed mixed graph, but which does not fall under the ancestral graph family. Table 1 summarises the main features of these relevant algorithms.

| Algorithm | Class | Discrete /Continuous data | Output | Intervention type | Data set |
|---|---|---|---|---|---|
| COmbINE | Constraint-based | Both | PAG | Perfect | Separate |
| HEJ | Constraint-based | Both | Cyclic Directed Mixed Graph | Perfect | Separate |
| JCI | Constraint-based | Both | Acyclic Directed Mixed Graph | Perfect/ Imperfect/ Uncertain | Pooled |
| ACI | Constraint-based | Both | Ancestral graph | Perfect/ Imperfect | Separate |
| mFGS-BS (This work) | Hybrid | Discrete | PAG | Perfect | Separate |

**Table 1**. Overview of relevant structure discovery algorithms that assume causal insufficiency and learn graphs from multiple interventions.

In this paper, we propose a novel hybrid structure learning algorithm called mFGS-BS, that produces a PAG from causally insufficient observational data and one or more interventional data sets. The paper is organised as follows: Section 2 provides preliminary information, Section 3 describes the proposed algorithm, Section 4 describes the evaluation process, Section 5 presents the empirical results, and we provide concluding remarks and discussions for future work in Section 6.

## 2. Preliminaries

The preliminaries focus on the methods relevant to the mFGS-BS algorithm that we later describe in Section 3. Specifically, subsection 2.1 covers ancestral graphs, subsection 2.2 covers interventions, subsection 2.3 covers the BDeu objective function, subsection 2.4 covers the Bayesian scoring method that assigns probabilities to CI tests, and subsection 2.5 covers the majority rule from mFCI.

### 2.1. Ancestral Graphs

Recall from Section 1 that a PAG represents a set of Markov equivalent MAGs, and that a MAG is an extended version of a DAG that represents relationships under the assumption of causal insufficiency. A MAG can contain the following types of edges: —, →, and ↔. The undirected edge A— B indicates that A is an ancestor of B or a selection variable, and B is an ancestor of A or a selection variable. The selection variable indicates the presence of selection bias in the data set. In this work, we will assume selection bias is not present in the data, and hence the undirected edge — will not be present in the MAGs or the PAGs we consider. Further, the directed edge A → B indicates parental or ancestral relationships, and the bidirected edge A ↔ B refers to the presence of a latent confounder where A and B are related but where neither A is an ancestor of B nor B is an ancestor of A. In a PAG, the variant mark (o) at the endpoint of edges indicates that the endpoint could be a tail (–) or an arrowhead (>) in the equivalence class of MAGs. For example, o→ in the PAG indicates that the edge can be either ↔ or

→ in the equivalent MAGs, whereas o—o indicates that the edge in the equivalent MAGs can be →, ← or ↔. Both MAGs and PAGs are acyclic graphs and do not allow the existence of almost directed cycles that may occur when A ↔ B is present and B is an ancestor of A (Richardson and Spirtes, 2000). Figure 1 illustrates an example of a DAG with latent variables $L_1$ and $L_2$, along with two examples of Markov equivalent MAGs that represent the conditional independencies between the observed variables in the marginal DAG and the latent variables, and the PAG representing the Markov equivalence class of those MAGs (Chobtham and Constantinou, 2020).

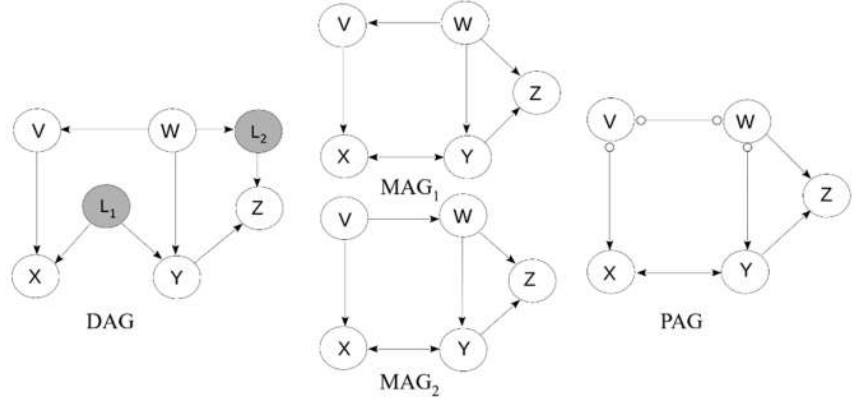

**Figure 1**. A causal DAG with observed variables $\{V, W, X, Y, Z\}$ ∪ latent variables $\{L_1, L_2\}$ in grey, with two examples of Markov equivalent MAGs, and the Markov equivalent PAG of MAGs.

*2.2. Interventions*

To resolve variant marks in a PAG requires that we look beyond observational data. As discussed in Section 1, interventional data can help us orientate some of these edges. Figure 2 illustrates the three different intervention mechanisms by comparing the pre-intervention and post-intervention actions. Specifically, a **Perfect intervention** is what Pearl describes as *do-calculus* (do(X)) where the intervened variable is set to a given state with no uncertainty (Pearl, 2000). A perfect intervention modifies the original causal structure by rendering the intervened variable independent of its causes (also referred to as *graph surgery*). On the other hand, an **Imperfect intervention** or a mechanism change (Tian and Pearl, 2001) can be viewed as having external intervention nodes that act like switching parents (I) on an intervened variable X for each external intervention node. Specifically, I = 1 activates the intervention where the target node X is parameterised over $\Theta_X^1$, whereas when I = 0 the intervention is deactivated and target node X is parameterised over $\Theta_X^0$ which would imply no external influence on node X. Applications of imperfect intervention are often observed in healthcare studies, where medicine and therapeutic actions often have an imperfect effect in terms of treating symptoms or curing diseases (Rickles, 2009). Lastly, an **Uncertain intervention** (Eaton and Murphy, 2007) represents the case where an external intervention I has multiple target nodes, or where the intervention on node X comes from more than one intervening route, as opposed to the imperfect intervention that assumes the relationship between intervention nodes and target nodes is one-to-one. Unlike perfect intervention, imperfect and uncertain interventions do not modify the graph and instead manipulate the node parameters.

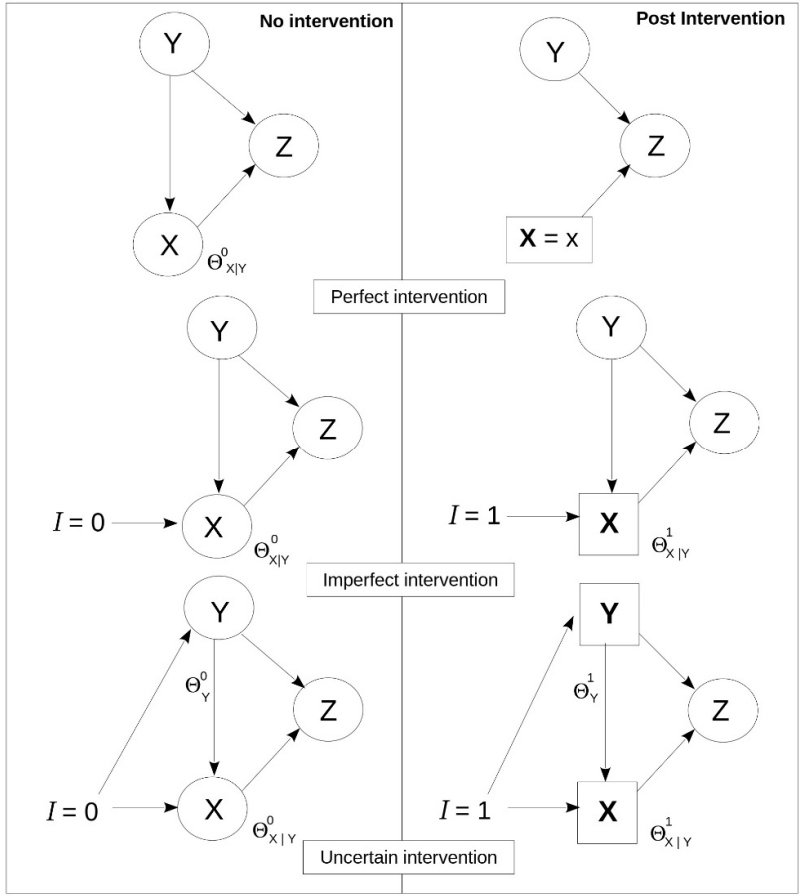

**Figure 2.** An illustration of the mechanisms of Perfect, Imperfect, and Uncertain interventions, where the square box represents the target node(s), $\Theta^0_{X|Y}, \Theta^0_Y$ are the parameters for nodes X and Y respectively when I = 0 (representing no intervention), and $\Theta^1_{X|Y}, \Theta^1_Y$ are the parameters for nodes X and Y respectively when I = 1 (representing an external imperfect or an uncertain intervention).

*2.3 The Bayesian Dirichlet equivalent uniform (BDeu) score*

Score-based algorithms use an objective function to assess each graph visited in the search space of graphs. The Bayesian Dirichlet equivalent uniform (BDeu) is one of the most commonly used objective functions in structure learning used to identify the maximum a posteriori (MAP) structure. It represents a variant of BD and BDe scores that assumes equivalent uniform priors. Importantly, these are decomposable scores where the total score of the graph represents the sum of the scores assigned to each of its nodes. A decomposable score is important for structure learning because most local scores can be reused, rather than recomputed, when exploring neighbouring graphs. BDeu is also score-equivalent in that it produces the same score for Markov equivalent structures, and hence it is used to search for the DAGs which entail the same joint probability distribution. The BD score was first introduced by Heckerman et al. (1995), under the assumption that the data follow a Dirichlet distribution. Pairing structure learning with BD as the objective function implies that the algorithm is searching for a DAG G that maximises the posterior probability P(G|D) given the data D. Structure learning from data can be viewed as an optimisation problem to maximise $P(G|D) \propto P(G) P(D|G)$ where the highest posterior probability of a learnt graph G is approximated to the highest log-likelihood score:

$$\log P(G|D) = \log P(G) + \log P(D|G)$$

where P(G) is the prior distribution over all DAGs. Because the search space of DAGs grows super-exponentially with the number of variables, it is impractical to specify informative priors for each DAG.

For simplicity, the prior distribution is often taken to be uniform. The BD score can be computed as follows:

$$P(D|G) = \prod_{i=1}^{N} \prod_{j=1}^{q_i} \left[ \frac{\Gamma(\Sigma_k \alpha_{ijk})}{\Gamma(\Sigma_k \alpha_{ijk} + \Sigma_k n_{ijk})} \prod_{k=1}^{|X_i|} \frac{\Gamma(\alpha_{ijk} + n_{ijk})}{\Gamma(\alpha_{ijk})} \right]$$

where $N$ is the number of variables, $q_i$ is the number of possible combinations of values of the parents of node $X_i$ (it is 1 if there is no parent), $j$ is the index over the combinations of values of the parents of node $X_i$, $|X_i|$ is the number of states of node $X_i$, $k$ is the index over the possible values of node $X_i$, $\Gamma$ is the Gamma function, $n_{ijk}$ is the total number of instances in data $D$ where the parents of node $X_i$ have the $j^{th}$ combination of values, and $\alpha_{ij}$ is the prior for the *equivalent sample size* (*ess*) – also known as the *imaginary sample size* (*iss*). The prior parameters are set to $\alpha_{ijk} = \alpha/|X_i|q_i$. The study by Silander et al. (2012) suggests that reasonable values for hyperparameter are $\alpha \in [1,20]$ where larger $\alpha$ values tend to produce denser DAGs. Because the BDeu score is very small, it is preferred to take its log value and its closed form expression is:

$$\text{BDeu score} = \sum_{i=1}^{N} \sum_{j=1}^{q_i} \left[ \log \frac{\Gamma(\alpha/q_i)}{\Gamma(\alpha/q_i + \Sigma_k n_{ijk})} + \sum_{k=1}^{|X_i|} \log \frac{\Gamma(\alpha/|X_i|q_i + n_{ijk})}{\Gamma(\alpha/|X_i|q_i)} \right] \quad (1)$$

BN structure learning represents an NP-hard problem, which means that searching over all possible graphs is intractable. One solution to this problem is the use of heuristics such as Greedy search, but these approaches will often get stuck in a local optimum solution. Other score-based solutions include exact learning which guarantee to return the highest scoring graph, but these are restricted to a relatively small set of variables and are out of the score of this paper. BDeu is established as one of the most commonly used objective functions in hybrid and score-based structure learning, and algorithms such as the hybrid GFCI and RFCI-BSC, as well as score-based Greedy Equivalent Search (GES) (Chickering, 2003), including its more efficient variant Fast Greedy equivalent Search (FGS) (Ramsey, 2015), use BDeu to greedily traverse the search space of graphs.

*2.4. Assigning probabilities to conditional independence tests and directed edges*

Previous works that assumed prior probabilities for the existence of directed edges, as opposed to a binary outcome, include those by Castelo and Siebes (2000) who introduced the idea of assigning subjective prior probabilities (specified by experts) to directed edges, and by Scutari (2017) who assumed the marginal uniform prior probabilities of directed edges A → B and A ← B to be ¼, while the prior probability of the independency between A and B to be ½ in a variant of the BD score called the Bayesian Dirichlet sparse score (BDs).

Hyttinen et al. (2014) proposed a Bayesian scoring method that applies prior probabilistic weights to the results obtained from CI tests. These prior probabilities are subjective and obtained from knowledge. In this paper, we modify this method so that the prior probabilities are objectively calculated from data, and are assigned to directed edges rather than to the results obtained from CI tests. These details are discussed in subsections 3.1 and 3.2. With reference to the method by Hyttinen et al. (2014), the posterior probability of CI ($P(r|D_{OBS})$), given observational data, is:

$$P(r|D_{OBS}) = \frac{\text{prior} \times P(D_{OBS}|r)}{\text{prior} \times P(D_{OBS}|r) + (1 - \text{prior}) \times P(D_{OBS}|\bar{r})} \quad (2)$$

where $r$ is an arbitrary CI that A and B are independent given $\mathbf{Z}$ (A ⊥ B | $\mathbf{Z}$), $\bar{r}$ is an arbitrary conditional dependence that A and B are dependent given $\mathbf{Z}$ (A ⊥̸ B | $\mathbf{Z}$), $\mathbf{Z}$ is the set of variables that is the separation set (Sepset) of variables A and B, prior is an informative or uninformative probability from knowledge that A ⊥ B | $\mathbf{Z}$ is true, $P(D_{OBS}|r)$ is the network score of A ⊥ B | $\mathbf{Z}$ (marginal likelihood), and $P(D_{OBS}|\bar{r})$ is the network score of A ⊥̸ B | $\mathbf{Z}$ (A → B or A ← B).

Similarly, Jabarri et al. (2017) used the BDeu score to obtain a posterior probability for CI in the hybrid RFCI-BSC algorithm, and assumed a uniform prior as the uninformative probability for each result obtained from CI tests as follows:

$$P(r|D_{OBS}) = \frac{P(D_{OBS}|r)}{P(D_{OBS}|r) + P(D_{OBS}|\bar{r})}$$

where $P(D_{OBS}|r)$ is the BDeu score (marginal likelihood) of structure $A \leftarrow Z \rightarrow B$ ($A \perp B | Z$), and $P(D_{OBS}|\bar{r})$ is the BDeu score of structure $A \leftarrow Z \rightarrow B$ and $A \rightarrow B$ ($A \not\perp B | Z$), and all variables in $Z$ are parents of both $A$ and $B$. These structures are proposed by Jabarri et al. (2017; 2020) to be the representation of all possible structures that correspond to the relevant CI tests. Since the marginal likelihoods can be found in the objective scores computed by score-based learning (Margaritis, 2005), the BDeu score of these structures can be used to derive the marginal likelihoods for discrete variables. The RFCI-BSC algorithm learns a structure from discrete observational data under the assumption of causal insufficiency. Moreover, it generates multiple PAGs by sampling over the joint posterior probabilities of CI, and picks the PAG with the highest joint posterior probability of CI. Since the decision for CI is determined with a random threshold, this makes the output of the algorithm nondeterministic. Empirical experiments show that RFCI-BSC fails to generate results for input data with sample size 10k or higher (Constantinou et al., 2021).

*2.5 majority rule FCI*

Recall from Section 1 that cFCI and mFCI are constraint-based algorithms and both represent extensions of FCI that improve edge orientation accuracy. cFCI is similar to cPC (Ramsey et al., 2012), where cPC does not (and cFCI does) assume causal insufficiency. In this paper, we also implement the majority rule from mFCI, in addition to the Bayesian scoring method described in subsection 2.4 to compute the prior probabilities of directed edges, to determine the likelihood of an unshielded triple being a v-structure (details will be in subsection 3.2.2).

Compared to FCI, cFCI performs additional CI tests on A and C given on all subsets of all neighbours of A and C including B, for each unshielded triple A-B-C, to more conservatively determine v-structures and orientate edges. This implies that cFCI discovers fewer, although with higher certainty, directed edges compared to FCI. For discrete data, the $G^2$ test can be used as the statistical test for determining CI between A and C conditional on B:

$$G^2 = 2 \sum_{a,c,b} n_{acb} \ln \frac{n_{acb} n_b}{n_{ab} n_{cb}}$$

where $n_{acb}$ is the total number of instances in data which $A = a, B = b$ and $C = c$. The calculation of the total number of instances of $n_{ab}, n_{cb}$ and $n_b$ is analogous to that of $n_{acb}$. A p-value associated with each statistical test result is then used to reject or accept CI, where a cut-off threshold of 0.05 is generally used establishing independence. For each unshielded triple A-B-C in the v-structure phase, the conservative rule from cFCI classifies each unshielded triple as either a definite v-structure, a definite non v-structure, or an ambiguous triple given the Sepsets, e.g. if B is not in any Sepsets A and C, the conservative rule will classify the unshielded triple A-B-C as a definite v-structure. Later, Colombo and Maathuis (2014) found that the conservative rule was orientating few of the v-structures and proposed the majority rule which can be viewed as the relaxed version of the conservative rule. They called this new variant the majority rule FCI (mFCI). Specifically, in mFCI, the majority rule classifies each unshielded triple A-B-C as:

   a. A v-structure if B is in less than 50% of the Sepsets of A and C,
   b. A non v-structure if B is in more than 50% of the Sepsets of A and C,
   c. An ambiguous triple if B is in 50% of the Sepsets of A and C.

## 3. The mFGS-BS algorithm

Recall from Section 2 that both the score-based and constraint-based algorithms covered in this paper produce a graph in the Markov equivalence class. This means that not all edges can be orientated given observational data, and only a few of the algorithms in the literature consider both observational and interventional data in an attempt to orientate as many edges as possible.

The mFGS-BS algorithm described in this section learns a PAG from both observational and interventional data, under the assumption of causal insufficiency and that the intervened variables are subject to perfect intervention. The novelty of mFGS-BS involves assigning probabilities to each possible directed edge. If the two opposing directions between a pair of variables both have probabilities that are higher than a given threshold, then a bidirected edge is assumed.

We first describe in subsection 3.1 how the probabilities of directed edges from a single observational data set can be obtained, and then describe in subsection 3.2 how we extend this concept to cases in which we want to learn a structure from both observational and interventional data. Subsection 3.3 provides the overall description of mFGS-BS.

### *3.1 Determining the probabilities of directed edges from a single observational data set*

We devise a new method to determine directed edges that is largely based on the methods of Hyttinen et al. (2014) and Jabbari et al. (2017) that focus on assigning probabilities to each result obtained from CI tests, which we previously covered in subsection 2.4. For the rest of this paper, we label observational data as $D_{OBS}$ and interventional data as $D_{INT}$. When assuming the unconditional independence between two variables A and B, we modify Equation (2) to consider the possibility of edges A B (i.e. no edge between A and B ), $A \rightarrow B$ and $A \leftarrow B$ in a DAG as follows:

$$P(A\ B|D_{OBS}) = \frac{P(A\ B) \times P(D_{OBS}|A\ B)}{P(A\ B) \times P(D_{OBS}|A\ B) + P(A \rightarrow B) \times P(D_{OBS}|A \rightarrow B) + P(A \leftarrow B) \times P(D_{OBS}|A \leftarrow B)}$$

Since $P(A\ B|D_{OBS}) + P(A \rightarrow B|D_{OBS}) + P(A \leftarrow B|D_{OBS}) = 1$ and $P(A\ B) + P(A \rightarrow B) + P(A \leftarrow B) = 1$, then:

$$1 - (P(A \rightarrow B|D_{OBS}) + P(A \leftarrow B|D_{OBS})) = \frac{(1 - (P(A \rightarrow B) + P(A \leftarrow B))) \times P(D_{OBS}|A\ B)}{(1 - (P(A \rightarrow B) + P(A \leftarrow B))) \times P(D_{OBS}|A\ B) + P(A \rightarrow B) \times P(D_{OBS}|A \rightarrow B) + P(A \leftarrow B) \times P(D_{OBS}|A \leftarrow B)} \quad (3)$$

where $P(A \rightarrow B)$ is the prior probability of directed edge $A \rightarrow B$, $P(A \leftarrow B)$ is the prior probability of directed edge $A \leftarrow B$ that we later describe in subsection 3.2, $P(D_{OBS}|A \rightarrow B)$ is the BDeu score of structure $A \rightarrow B$, and $P(D_{OBS}|A \leftarrow B)$ is the BDeu score of structure $A \leftarrow B$.

Because we assume that the learnt ancestral graph is a PAG that may contain bidirected edges, the bidirected edge $A \leftrightarrow B$ corresponds to the dependency between A and B from the assumed true structure $A \leftarrow L \rightarrow B$ ($A \not\perp B$) where L is a latent confounder. The dependency between A and B in a PAG can be $A \rightarrow B$, $A \leftarrow B$ or $A \leftrightarrow B$. Because Equation (3) is not suitable to calculate the posterior probabilities of these types of edges, we devise two equations: (1) calculating $P(A \rightarrow B|D_{OBS})$ by ignoring $A \leftarrow B$, as described in Case 1 below, and (2) calculating $P(A \leftarrow B|D_{OBS})$ by ignoring $A \rightarrow B$, as described in Case 2 below. These enable us to calculate the probabilities of each of these directed edges independently. If the posterior probabilities of both directed edges $A \rightarrow B$ and $A \leftarrow B$ are higher than a given threshold, then mFGS-BS is not be able to orientate the given directed edges and will produce the bidirected edge $A \leftrightarrow B$.

**Case 1**: Calculate $P(A \rightarrow B|D_{OBS})$ given the assumption that $P(A \leftarrow B|D_{OBS}) = 0$, $P(D_{OBS}|A \leftarrow B) = 0$ and $P(A \leftarrow B) = 0$ from Equation (3), then:

$$1 - P(A \rightarrow B|D_{OBS}) = \frac{(1 - P(A \rightarrow B)) \times P(D_{OBS}|A\ B)}{(1 - P(A \rightarrow B)) \times P(D_{OBS}|A\ B) + P(A \rightarrow B) \times P(D_{OBS}|A \rightarrow B)}$$

**Case 2:** Calculate $P(A \leftarrow B|D_{OBS})$ given the assumption that $P(A \rightarrow B|D_{OBS}) = 0$, $P(D_{OBS}|A \rightarrow B) = 0$ and $P(A \rightarrow B) = 0$ from Equation (3), then:

$$1 - P(A \leftarrow B|D_{OBS}) = \frac{(1 - P(A \leftarrow B)) \times P(D_{OBS}|A\ B)}{(1 - P(A \leftarrow B)) \times P(D_{OBS}|A\ B) + P(A \leftarrow B) \times P(D_{OBS}|A \leftarrow B)}$$

From this, we define the posterior probabilities of directed edges as specified by Definition 1.

**Definition 1** Assuming the learnt graph is a PAG, we define a bidirected edge $A \leftrightarrow B$ as the dependency between A and B derived from the possibility of both $A \rightarrow B$ and $A \leftarrow B$, where the posterior probabilities $P(A \rightarrow B|D_{OBS})$ and $P(A \leftarrow B|D_{OBS})$ are:

$$P(A \rightarrow B|D_{OBS}) = 1 - \frac{(1 - P(A \rightarrow B)) \times P(D_{OBS}|A\ B)}{(1 - P(A \rightarrow B)) \times P(D_{OBS}|A\ B) + P(A \rightarrow B) \times P(D_{OBS}|A \rightarrow B)}$$

$$P(A \leftarrow B|D_{OBS}) = 1 - \frac{(1 - P(A \leftarrow B)) \times P(D_{OBS}|A\ B)}{(1 - P(A \leftarrow B)) \times P(D_{OBS}|A\ B) + P(A \leftarrow B) \times P(D_{OBS}|A \leftarrow B)}$$

### 3.2 Determining the probabilities of directed edges from both observational and interventional data sets

We extend the approach above to learn from an observational data set and one or more interventional data sets, which the algorithm processes in turn. For each interventional data set, $INT_i$, the algorithm uses Equations (4) and (5) to determine the posterior probability of each directed edge. We use the term "posterior" here to reflect the fact that this probability, denoted for example, $P(A \rightarrow B|D_{INT_i})$, is based **both** on the current interventional data set being processed **and** all previous data sets processed.

$$P(A \rightarrow B|D_{INT_i}) = 1 - \frac{(1-P(A \rightarrow B)) \times P(D_{INT_i}|A\ B)}{(1-P(A \rightarrow B)) \times P(D_{INT_i}|A\ B) + P(A \rightarrow B) \times P(D_{INT_i}|A \rightarrow B)} \quad (4)$$

$$P(A \leftarrow B|D_{INT_i}) = 1 - \frac{(1-P(A \leftarrow B)) \times P(D_{INT_i}|A\ B)}{(1-P(A \leftarrow B)) \times P(D_{INT_i}|A\ B) + P(A \leftarrow B) \times P(D_{INT_i}|A \leftarrow B)} \quad (5)$$

The term $P(A \rightarrow B)$ on the right hand side of Equation (4) represents the objective prior probability of directed edge $A \rightarrow B$ based on the previously processed data sets. The term $P(A \leftarrow B)$ plays an analogous role as the objective prior for $A \leftarrow B$ in Equation (5). The prior for $A \rightarrow B$ is taken to be **either** the posterior for that directed edge computed in the previous iteration, that is, $P(A \rightarrow B|D_{INT_{i-1}})$, **or** a prior derived using Equation (6) **whichever is the larger**.

$$P(A \rightarrow B) = \max\{P_{FGS}(A \rightarrow B)|D_{OBS,INT_{1:i-1}}, P(A \rightarrow B)_{A \rightarrow B \leftarrow C}|D_{OBS}\}$$
$$+ \sum_{k=1}^{i-1} P(A - B)_{\text{local BDeu of B,target }=A}|D_{OBS,INT_k} \quad (6)$$

where $P(A \rightarrow B)$ is computed from three factors on the right hand side of Equation (6):

- Factor 1: $P_{FGS}(A \rightarrow B)|D_{OBS,INT_{1:i-1}}$ is the probability of directed edge $A \rightarrow B$ over all previously learnt CPDAGs from FGS across $D_{OBS,INT_{1:i-1}}$ (further details are provided in subsection 3.2.1).
- Factor 2: $P(A \rightarrow B)_{A \rightarrow B \leftarrow C}|D_{OBS}$ is the probability of directed edge $A \rightarrow B$ calculated from the ratio of Sepsets determining v-structure $A \rightarrow B \leftarrow C$ using the majority rule from $D_{OBS}$ (further details are provided in subsection 3.2.2).
- Factor 3: $\sum_{k=1}^{i-1} P(A - B)_{\text{local BDeu of B,target }=A}|D_{OBS,INT_k}$ is the summation of all relative changes in the local BDeu scores of node B compared to $D_{OBS}$, when the intervened variable is A across all previously learnt $D_{INT}$. The relative changes in the local BDeu scores are described in subsection 3.2.3.

#### 3.2.1 Factor 1: Determining the probabilities of directed edges given the occurrence rates of each directed edge over all learnt CPDAGs

The first, out of the three, factors used to calculate the prior probability of a directed edge is based on the occurrence rate of each directed edge derived from the probability of directed edge $A \rightarrow B$ ($P_{FGS}(A \rightarrow B)|D_{OBS,INT_{1:i-1}}$) over all learnt CPDAGs obtained by applying FGS to each input data set. Specifically, :

$$P_{FGS}(A \rightarrow B)|D_{OBS,INT_{1:i-1}} = \frac{\#\text{directed edge}(A \rightarrow B)}{\#\text{total directed edge}(A \rightarrow B) + \#\text{total directed edge}(A \leftarrow B)}$$

where:

$$\text{directed edge}(A \rightarrow B) = \begin{cases} 1 & : \text{if } A \rightarrow B \text{ is in a learnt CPDAG} \\ 0.5 & : \text{if } A - B \text{ is in a learnt CPDAG and the intervened variable} = A \\ 0 & : \text{otherwise} \end{cases}$$

and:

$$\text{total directed edge}(A \rightarrow B) = \begin{cases} 1 : \text{if } A \rightarrow B \text{ is in a learnt CPDAG} \\ 1 : \text{if } A - B \text{ is in a learnt CPDAG and the intervened variable} = A \\ 0 : \text{otherwise} \end{cases}$$

The total number of directed edges $A \rightarrow B$ represents the number of directed edges $A \rightarrow B$ present in each of the learnt CPDAGs. Note that CPDAGs learnt from interventional data should not produce directed edges entering the intervened variable due to the graph surgery mechanisms illustrated in Figure 2 (i.e., interventions are rendered independent of their parents). For example, if the undirected edge $A - B$ is present in the learnt CPDAG when we intervene on node A, the algorithm assigns probability 0 for directed edge $A \leftarrow B$ and probability 0.5 for directed edge $A \rightarrow B$ to account for the risk of false positive edges learnt by FGS, since it does not produce bidirected edges in the presence of latent confounders (Ogarrio et al., 2016).

It is important to clarify that in the absence of intervention, an undirected edge in the learnt CPDAG does not imply equal probability for either direction (Kummerfeld, 2021). The correct probability for each directed edge can be obtained by enumerating all possible DAGs from the learnt CPDAG. However, this tends to increase the computational complexity of the algorithm substantially, especially in the case of mFGS-BS which is designed to produce a CPDAG for each input data set. For simplicity and reasons of efficiency, when an undirected edge is present in a learnt CPDAG, mFGS-BS assumes a probability of 0.5 for either direction.

### 3.2.2 Factor 2: Determining the probabilities of directed edges given the ratios of Sepsets determining v-structures

Because the joint probability distribution from interventional data will not capture all dependencies, we consider the v-structures as determined by observational data. Therefore, interventional data is not used by this factor. In mFCI, the v-structures are obtained from unshielded triples that are part of an initial undirected graph determined by statistical CI tests. Then, the majority rule in mFCI is used to definitively orientate the edges of unshielded triples A-B-C into v-structures $A \rightarrow B \leftarrow C$, determined by the ratio of Sepsets (Colombo and Maathuis, 2014). In this paper, we use a novel method to instead calculate the probabilities of these directed edges, where $P(A \rightarrow B)_{A \rightarrow B \leftarrow C}|D_{OBS}$ and $P(C \rightarrow B)_{A \rightarrow B \leftarrow C}|D_{OBS}$ correspond to the individual probabilities of directed edges $A \rightarrow B$ and $C \rightarrow B$ in producing v-structure $A \rightarrow B \leftarrow C$ given the observational data. In order to assign a probability to directed edges in an unshielded triple A-B-C, mFGS-BS considers how many of the Sepsets of A and C contain B. If B is in less than 50% of the Sepsets of A and C (i.e., the ratio of Sepsets < 0.5) then we assume that B does not block an active path between A and C. Hence, the likelihood of v-structure $A \rightarrow B \leftarrow C$ will be higher than 0.5, and from this we deduce that $P(A \rightarrow B)_{A \rightarrow B \leftarrow C}|D_{OBS} > 0.5$ and $P(C \rightarrow B)_{A \rightarrow B \leftarrow C}|D_{OBS} > 0.5$. Conversely, if B is in $\geq 50\%$ of the Sepsets of A and C, we deduce that the unshielded triple A-B-C is unlikely to be a v-structure and that instead is likely to be either $A \rightarrow B \rightarrow C$, $A \leftarrow B \rightarrow C$ or $A \leftarrow B \leftarrow C$. These assumptions lead to Equation (7) and (8) which are calculated independently as follows:

$$P(A \rightarrow B)_{A \rightarrow B \leftarrow C}|D_{OBS} = P(C \rightarrow B)_{A \rightarrow B \leftarrow C}|D_{OBS} = \begin{cases} 1-\text{the ratio of Sepset} & : \text{the ratio of Sepset} < 0.5 \\ 0.5 & : \text{the ratio of Sepset} \geq 0.5 \end{cases} \quad (7)$$

$$P(A \leftarrow B)_{A \rightarrow B \leftarrow C}|D_{OBS} = P(C \leftarrow B)_{A \rightarrow B \leftarrow C}|D_{OBS} = 0.5 \quad (8)$$

where the ratio of Sepsets $= \frac{|\text{Sepsets of A and C which contain B}|}{|\text{all Sepsets of A and C}|}$, |Sepsets of A and C which contain B| and |all Sepsets of A and C| represent the number of Sepsets in $D_{OBS}$. $P(A \rightarrow B)_{A \rightarrow B \leftarrow C}|D_{OBS}$, $P(C \rightarrow B)_{A \rightarrow B \leftarrow C}|D_{OBS}$ from Equation (7), $P(A \leftarrow B)_{A \rightarrow B \leftarrow C}|D_{OBS}$ and $P(C \leftarrow B)_{A \rightarrow B \leftarrow C}|D_{OBS}$ from Equation (8) are assigned the value of 0.5 for the reasons covered in subsection 3.2.1.

*3.2.3 Factor 3: Determining the probability of directed edges given the relative changes in local BDeu scores*

From Equation (1), we know the BDeu score of a graph represents the summation of all local BDeu scores assigned to each node within that graph. The local BDeu score for node i ($Z_i$) (Cussens, 2012) is denoted as:

$$Z_i = \sum_{j=1}^{q_i} \left[ \log \frac{\Gamma(\alpha/q_i)}{\Gamma(\alpha/q_i + \Sigma_k n_{ijk})} + \sum_{k=1}^{|X_i|} \log \frac{\Gamma(\alpha/|X_i|q_i + n_{ijk})}{\Gamma(\alpha/|X_i|q_i)} \right]$$

The effect of an intervention represents the difference between pre and post-intervention distributions of the children of a target node (Zhang, 2006). We consider the difference in their local BDeu scores to represent the effect of the intervention, assuming the sample size of the input observational data is the same with the sample size of the interventional data when computing this difference. From this, we obtain the relative change in the local BDeu scores as described by Definition 2.

**Definition 2** Assuming equal sample size for both observational and interventional data, the relative change in the local BDeu scores between pre-intervention ($Z_i|D_{OBS}$) and post-intervention ($Z_i|D_{INT}$) of node i is:

$$\left| \frac{Z_i|D_{OBS} - Z_i|D_{INT}}{Z_i|D_{OBS}} \right| \quad (9)$$

For example, when we intervene on node A when $A \rightarrow B$ is present in the graph, then we would expect the effect of this intervention to be reflected in the probability distribution of B. When A is the intervened variable and the undirected edge $A - B$ is learnt by FGS given $D_{INT}$, we are interested in the likelihood of the directed edge $A \rightarrow B$ being present in the true graph. In this case, the probability of directed edge $A \rightarrow B$ is measured by Factor 3 in terms of the relative change in the local BDeu score of node B, given $D_{INT}$ and $D_{OBS}$, as defined by Equation (9).

**Example 1.** This example is described with reference to Figure 3, and assumes that the true DAG is the one shown in Figure 1. Figure 3a shows the undirected graph as constructed by the CI tests given $D_{OBS}$, to determine unshielded triples. Figures 3b, 3c and 3d present the three hypothetical CPDAGs learnt by FGS from three different data sets. We first illustrate how to derive Factor 2 in Table 2, where the first column shows that the CI tests over V and Y, given the unshielded triple $V - X - Y$ in Figure 3a, return 3 Sepsets with p-values greater than the cut-off threshold of 0.05. The only Sepset of node V and Y that contains X is {W, X, Z}. This means that the ratio of Sepsets in determining the given v-structure will be 0.333, as shown in the second column in Table 2. The third and fourth columns show how we arrive at the calculation of Factor 2, given Equations (7) and (8) respectively, each of which corresponds to a probability of the directed edge being present in the true graph.

| Sepsets of V and Y | The ratio of Sepsets containing X | Factor 2: $P(V \rightarrow X)_{V \rightarrow X \leftarrow Y}|D_{OBS}$ $P(Y \rightarrow X)_{V \rightarrow X \leftarrow Y}|D_{OBS}$ given Equation (7) | Factor 2: $P(V \leftarrow X)_{V \rightarrow X \leftarrow Y}|D_{OBS}$ $P(Y \leftarrow X)_{V \rightarrow X \leftarrow Y}|D_{OBS}$ given Equation (8) |
|---|---|---|---|
| {W} {W, X, Z} {Z} | $\frac{1}{3} = 0.333$ | 1-0.333 = 0.667 | 0.5 |

**Table 2.** How the probabilities of directed edges of Factor 2 are calculated, given the unshielded triple $V - X - Y$ in Example 1 and with reference to Figure 3a.

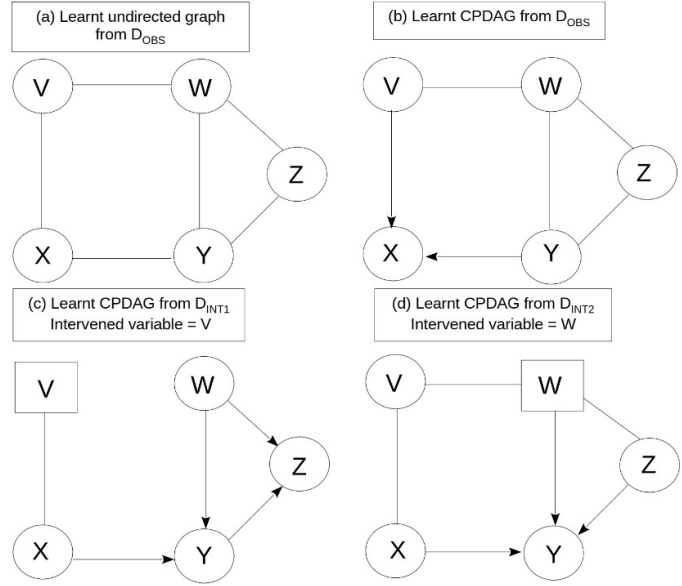

**Figure 3.** (a) The undirected graph produced by the CI tests given $D_{OBS}$, (b)-(d) and the three CPDAGs learnt by FGS from observational and interventional data ($D_{OBS}$, $D_{INT_1}$ and $D_{INT_2}$) generated based on the DAG shown in Figure 1, with variables targeted for intervention $T_1=\{V\}$, $T_2=\{W\}$ shown in rectangles.

Table 3 illustrates how Factor 3 is calculated, that produces the relative change in the local BDeu scores as described in subsection 3.2.3. The example is based on one observational data set, two interventional data sets, and one intervened variable per interventional data set as shown in Figure 3c and Figure 3d. Figure 3c shows that the undirected edge $V - X$ is learnt by FGS given $D_{INT_1}$. When V is the intervened variable, we observe that the relative change in the local BDeu score of node X is 0.0119 from the effect of this intervention, so this increases the probability of directed edge $V \to X$ being present in the true graph. Table 3 also shows the relative changes in the local BDeu score of V and Z are 0.0174 and 0.0001 respectively when W is the intervened variable in Figure 3d.

| Directed edges | Interventional data sets | Intervened variables | Local BDeu score (pre-intervention) | Local BDeu score (post-intervention) | Relative change in local BDeu scores given Equation (9) |
|---|---|---|---|---|---|
| $V \to X$ | $D_{INT_1}$ | V | X = -11507 | X = -11370 | 0.0119 |
| $W \to V$ | $D_{INT_2}$ | W | V = -14274 | V = -14026 | 0.0174 |
| $W \to Z$ | $D_{INT_2}$ | W | Z = -6936 | Z = -6935 | 0.0001 |

**Table 3.** An example of calculating the relative change in the local BDeu scores as described in Example 1 and with reference to Figure 3c and Figure 3d.

Finally, Table 4 presents the outputs produced by each of the three factors, and with reference to the directed edges presented in the first column. The calculations in the second, third and fourth columns correspond to the outputs of Factors 1, 2 and 3 respectively. In calculating Factor 1 for directed edge $X \to Y$, Figure 3b, 3c and 3d show that $X \leftarrow Y$ appears once and $X \to Y$ appears twice across the three CPDAGs, thus $P_{FGS}(X \to Y)|D_{OBS,INT_{1:2}} = 0.67$. For directed edge $W \to V$, Figure 3b shows $W - V$, Figure 3c shows no edge, and Figure 3d shows $W - V$ given $D_{INT_2}$ and hence, $P_{FGS}(V \to W)|D_{OBS,INT_{1:2}}$ is set to 0 and $P_{FGS}(W \to V)|D_{OBS,INT_{1:2}}$ to 0.5. This is because W is the intervened variable in Figure 3d, and from this we can conclude that if an edge is discovered between V and W, then the direction of that edge can only be entering V. Note that $P_{FGS}(W \to V)|D_{OBS,INT_{1:2}}$ is set to 0.5 and not to 1 because FGS suggests $W - V$ instead of $W \to V$. Finally, the fifth column of Table 4 shows the overall calculation for the prior probability of each directed edge, that takes into consideration all three factors, given Equation (6).

| Directed edges | Factor 1: $P_{FGS}(A \rightarrow B)\|D_{OBS,INT_{1:2}}$ | Factor 2: $P(A \rightarrow B)_{A \rightarrow B \leftarrow C}\|D_{OBS}$ | Factor 3: $\sum_{k=1}^{2} P(A-B)_{\text{local BDeu of B,target}=A}\|D_{OBS,INT_k}$ | $P(A \rightarrow B)$ given Equation (6) |
|---|---|---|---|---|
| X → Y | 0.67 | 0.5 | - | 0.67 |
| Y → X | 0.34 | 0.67 | - | 0.67 |
| V → X | 0.75 | 0.67 | 0.0119 | 0.7619 |
| W → V | 0.5 | 0 | 0.0174 | 0.5174 |
| V → W | 0 | 0 | - | 0 |
| W → Y | 1 | 0 | - | 1 |
| W → Z | 0.75 | 0 | 0.0001 | 0.7501 |

**Table 4.** Examples of the calculation of the prior probability of directed edges with reference to Example 1, Figure 3, Table 2 and Table 3.

### 3.3 Algorithm mFGS-BS

We now use the concepts described in subsections 3.1 and 3.2 to formulate the mFGS-BS algorithm. The pseudocode of mFGS-BS is provided in Algorithm 1. The algorithm takes as an input an observational data set and one or more interventional data sets, the set of variables targeted for intervention for each interventional data, and the hyperparameters specified in Algorithm 1. The overall process of mFGS-BS is shown in Figure 4. The first step in Algorithm 1 performs CI tests given an observational data set. Steps 2 to 4 derive the initial prior probabilities of directed edges forming v-structures and the probabilities of directed edges learnt by FGS given an observational data set. Step 5 then iteratively calculates the posterior probabilities of directed edges derived from each interventional data set, as described in subsection 3.2. In the last steps, a PAG is constructed from the posterior probabilities of directed edges obtained after processing the last interventional data set, based on a hyperparameter cut-off threshold used to determine the existence of a directed edge or bidirected edge.

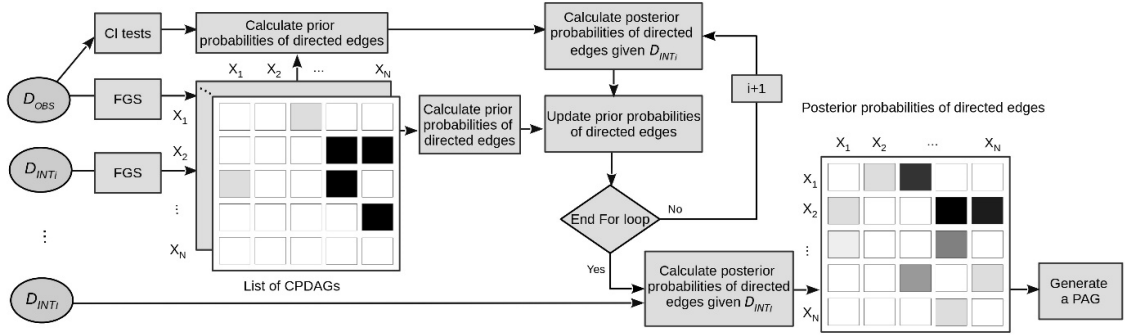

**Figure 4.** The overall process of the mFGS-BS algorithm that iteratively processes data sets and calculates posterior probabilities of directed edges to generate a PAG.

| | |
|---|---|
|Algorithm 1: mFGS-BS (majority rule and Fast Greedy equivalent Search with Bayesian Scoring)| |

**Input:** interventional data sets $D_{INT_I}$, an observational data $D_{OBS}$, intervened variable sets $T_I$, significance threshold t, posterior probability cut-off threshold c, maximum Sepset size k
**Output:** a PAG

Step 1  Set up a complete undirected graph $\mathcal{U}$ and Sepset **Z** size =0
**Repeat**
  Remove the dependencies for each pair (A, B) in $\mathcal{U}$ if they become independent given subsets of Sepset **Z**, determined by significance threshold t in $D_{OBS}$
  Sepset **Z** size = Sepset **Z** size +1
**Until** Sepset **Z** size k has been tested

Step 2  Given unshielded triple A − B − C from $\mathcal{U}$ resulting from Step 1, perform CI tests, with significance threshold t, on A and C given all neighbours of A and C including B, given $D_{OBS}$ and calculate Factor 2 $P(A \rightarrow B)_{A \rightarrow B \leftarrow C}|D_{OBS}$, $P(A \leftarrow B)_{A \rightarrow B \leftarrow C}|D_{OBS}$, $P(C \rightarrow B)_{A \rightarrow B \leftarrow C}|D_{OBS}$ and $P(C \leftarrow B)_{A \rightarrow B \leftarrow C}|D_{OBS}$ according to Equations (7) and (8)

Step 3  Run FGS on $D_{OBS}$ and add the learnt CPDAG to the list $\mathcal{L}_G$

Step 4  **For each** pair (A, B) over all variables
  Calculate the prior probabilities $P(A \rightarrow B), P(A \leftarrow B)$ for each possible directed edge $A \rightarrow B, A \leftarrow B$
    where $P(A \rightarrow B) = \max\{P_{FGS}(A \rightarrow B)|D_{OBS}, P(A \rightarrow B)_{A \rightarrow B \leftarrow C}|D_{OBS}\}$
  Calculate the posterior probabilities $P(A \rightarrow B|D_{INT_1}), P(A \leftarrow B|D_{INT_1})$ for each possible directed edge $A \rightarrow B, A \leftarrow B$ according to Equations (4) and (5)
**End for**

Step 5  **For** i=1 to I-1
  Run FGS on $D_{INT_i}$ and add the learnt CPDAG to the list $\mathcal{L}_G$
  **For each** pair (A, B) over all variables
    **If** A is the intervened variable
      Calculate Factor 3 $P(A - B)_{local\ BDeu\ of\ B, target=A}|D_{OBS,INT_i}$ given Equation (9) and add it to the list $\mathcal{L}_S$
    **End if**
    Calculate the prior probabilities $P(A \rightarrow B), P(A \leftarrow B)$ for each possible directed edge $A \rightarrow B$ and $A \leftarrow B$ given Equation (6), where Factor 1 is calculated given $\mathcal{L}_G$, Factor 2 is calculated given Equations (7) and (8) in Step 2, and Factor 3 is calculated given $\mathcal{L}_S$
    $P(A \rightarrow B) \leftarrow \max\{P(A \rightarrow B), P(A \rightarrow B|D_{INT_i})\}$
    $P(A \leftarrow B) \leftarrow \max\{P(A \leftarrow B), P(A \leftarrow B|D_{INT_i})\}$
    Calculate the posterior probabilities $P(A \rightarrow B|D_{INT_{i+1}}), P(A \leftarrow B|D_{INT_{i+1}})$ for each possible directed edge $A \rightarrow B, A \leftarrow B$ according to Equations (4) and (5)
  **End for**
**End for**

Step 6  **Repeat until no cycles or an almost cyclic are present in the output graph**
  **For each** pair (A, B) in all variables
    Select edge $A \rightarrow B$ if the posterior probability $P(A \rightarrow B|D_{INT_I})$ is higher than threshold c;
    Select edge $A \leftrightarrow B$ if the posterior probabilities $P(A \rightarrow B|D_{INT_I})$ and $P(A \leftarrow B|D_{INT_I})$ are both higher than threshold c;
    Select edge $A \circ\!\!-\!\!\circ B$ if the posterior probabilities $P(A \rightarrow B|D_{INT_I})$ and $P(A \leftarrow B|D_{INT_I})$ are both lower or equal to threshold c, but A − B exists in $\mathcal{U}$.
  **End for**
  **If** the graph contains a cycle or an almost cyclic
    Remove an edge that causes a cycle, or an almost cycle, where the edge selected is the one that has the lowest posterior probability.
  **End If**
**End**

Step 7  Output a PAG by combining the set of edges learnt from Step 6

## 4. Case studies, data simulation and Evaluation

We consider six networks that greatly vary in dimensionality. All six case studies are based on real networks constructed by experts and are taken from the literature. These are: a) **Asia** which is a small network that captures the relationships between a visit to Asia, tuberculosis and lung cancer (Lauritzen and Spiegelhalter, 1988), b) **Sports** which is a small network that measures the effect of possession in football matches, on shots generated and goals scored (Constantinou et al., 2020), c) **Property** which is a medium-size network for investment decision making in the UK property market (Constantinou et al., 2020), d) **Alarm** which is a medium-size network of an alarm notification system for patients (Beinlich et al., 1989), e) **ForMed** which is a large network modelling the risk of violent reoffending in mentally ill prisoners (Constantinou et al., 2020), and f) **Pathfinder** which is a very large network for diagnosis of lymph-node diseases (Heckerman et al., 1992). The properties of these six networks are provided in Table 5.

| Network size | Network | Variables | Edges | Max in-degree | Free parameters |
|---|---|---|---|---|---|
| **Small** | Asia | 8 | 8 | 2 | 18 |
|  | Sports | 9 | 15 | 2 | 1,049 |
| **Medium** | Property | 27 | 31 | 3 | 3,056 |
|  | Alarm | 37 | 46 | 4 | 509 |
| **Large** | ForMed | 88 | 138 | 6 | 912 |
| **Very Large** | Pathfinder | 109 | 195 | 5 | 71,890 |

**Table 5.** The properties of the six real-world networks considered for evaluation.

We use the networks to generate one observational and up to 10 interventional data sets. The true MAGs and true DAGs for each of the networks are available in the Bayesys repository (Constantinou et al., 2020). For each true DAG, we consider observational and interventional data sets over two sample sizes (n=1k and n=10k). Interventional data are generated using the *bnlearn* R package (Scutari, 2019). For each data set, we randomly choose one or five variables to be targeted for intervention. This means it is possible for the same variable is targeted for intervention in more than one interventional data set. We remove all incoming edges entering intervened variables, and we assume a uniform distribution for each state of variables targeted for intervention, before the intervention is set, as in (Korb et al., 2004). Finally, 10% of the variables in the smaller networks (Asia and Sports) and 5% of the variables in the larger networks (Property, Alarm, Formed and Pathfinder) are made latent.

The structure learning performance is evaluated using the graphical measures of Precision, Recall, F1 and the Balance Scoring Function (BSF). The F1 score ranges from 0 to 1, and represents the harmonic mean of Precision and Recall, calculated as follows: $F1 = 2 \times \left(\frac{\text{Precision} \times \text{Recall}}{\text{Precision} + \text{Recall}}\right)$. The BSF score (Constantinou, 2019) considers all four confusion matrix parameters (TP, TN, FP and FN) to return a balanced score $\text{BSF} = 0.5 \times \left(\frac{TP}{a} + \frac{TN}{i} - \frac{FP}{i} - \frac{FN}{a}\right)$, where a is the number of edges in the true MAG, i is the number of independencies in the true MAG, $i = \frac{N(N-1)}{2} - a$, and N is the number of variables. The BSF score ranges from -1 to 1, where 1 corresponds to a perfect match between learnt and true graphs, 0 represents a score equivalent to that obtained from an ignorant empty or a fully connected graph, and -1 corresponds to the worst possible mismatch. To minimise uncertainty, we repeat the experiments five times per algorithm and obtain the average scores.

We compare the graphical scores obtained by mFGS-BS to those obtained by COmbINE, RFCI-BSC and GFCI, which are three similar algorithms that also produce a PAG. RFCI-BSC assigns probabilities to CIs that are used to learn a PAG, which is the most similar approach to mFGS-BS, whilst the well-establish GFCI supports latent variables and has been shown to more accurate than FCI and RFCI (Ogarrio et al., 2016). An important difference amongst these algorithms is that COmbINE enables learning from multiple interventional data sets while RFCI-BSC and GFCI do not. RFCI-BSC and GFCI are hybrid algorithms which assume the input data are observational. We therefore combined the observational and interventional data sets into a single data set, which we used as an input to these algorithms. This serves as a baseline experiment where the RFCI-BSC and GFCI algorithms produce a result given all data, but without taking advantage of interventional information.

COmbINE was tested using the MATLAB implementation by Triantafillou (2016) while RFCI-BSC and GFCI were tested using the *rcausal* package, which is the R wrapper for Tetrad Library (Wongchokprasitti, 2019). Note the output of COmbINE represents a special type of PAG that contains dashed edges (---) indicating uncertainty about the existence of an edge learnt from each interventional data set. Since we are interested in the direction of causation, all output PAGs are measured against the ground truth MAG using the penalty scores described in Table 6. Regarding the hyperparameter inputs of the algorithms, the significant threshold for the G-square hypothesis test is set to 0.05, and the max Sepset size of the conditioning set is set to 10, in all algorithms. The posterior probability cut-off threshold of mFGS-BS is set to 0.5, and the default *ess* of BDeu in mFGS-BS, RFCI-BSC and GFCI is set to 1. We also apply a runtime limit of four hours to each graph learnt/experiment for all algorithms.

| True edges | Predicted edges | Penalty | Result |
|---|---|---|---|
| A ↔ B | A → B, A ⇢ B, A ← B, A ⇠ B, A   B | 1 | True Positive = 0 |
| A → B | A ↔ B, A ← B, A ⇠ B, A   B | 1 | True Positive = 0 |
| A ← B | A ↔ B, A → B, A ⇢ B, A   B | 1 | True Positive = 0 |
| A   B | A → B, A ⇢ B, A ← B, A ⇠ B, A ↔ B | 1 | True Negative = 0 |
| A — B | A → B, A ⇢ B, A ← B, A ⇠ B, A ↔ B | 0.5 | True Positive = 0.5 |
| A → B | Ao— oB, Ao---oB | 0.5 | True Positive = 0.5 |
| A ← B | Ao— oB, Ao---oB | 0.5 | True Positive = 0.5 |
| A → B | A o→ B, A o⇢ B | 0.25 | True Positive = 0.75 |
| A ← B | A ←o B, A ⇠o B | 0.25 | True Positive = 0.75 |

**Table 6.** The edge and orientation penalty scores used by the scoring metrics, where ⇢ represents one of the output edges of COmbINE.

## 5. Empirical results

The results are separated into four subsections. We start with subsection 5.1, where we measure the sensitivity to the order of interventional data sets, we use the Alarm network to generate 5 and 10 interventional data sets with sample sizes 1k and 10k by intervening on a random single variable per data set and 5% of the variables in the data are made latent. Then, we randomise 20 orderings of 5 and 10 interventional data sets, and evaluate the results. In subsection 5.2, we assess the impact of each of the three factors described in subsection 3.2.2 on graphical learning accuracy. Subsection 5.3 compares the results of mFGS-BS to those of the other algorithms when we intervene on a single variable per interventional data set, and subsection 5.4 when we intervene on five variables per interventional data set.

*5.1 Assessing the sensitivity of the ordering of interventional data sets*

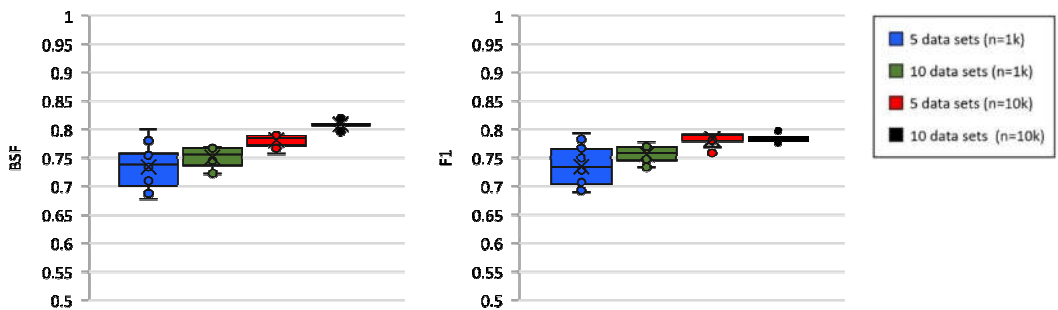

**Figure 5.** The boxplots show the BSF and F1 scores of mFGS-BS from 20 random interventional data orderings generated from the Alarm network, assuming one intervened variable and 5% latent variables per data set, over two sample sizes and two numbers of interventional data sets. The boxplots report the average values (the symbol x in the box) along with the median (the middle line of the box), and the maximum and minimum scores (the whiskers of the box).

The mFGS-BS algorithm updates the posterior probabilities of directed edges by taking into consideration a single interventional data set at a time. In this subsection, we evaluate how this ordering might influence the graphical performance of the algorithm. This experiment involves the different combinations of 5 and 10 interventional data sets, and sample sizes 1k and 10k. The boxplot in Figure 5 shows the BSF and F1 scores of mFGS-BS when applied to each hyperparameter setting involving the Alarm network. Each of the four scenarios involves 20 randomised orderings of interventional data. The results show that the average BSF score is 0.73±0.0363 when we have 5 interventional data sets at 1k sample size each, and the variability decreases to 0.81±0.0058 for 10 interventional data sets at 10 sample size each. We observe that the average F1 scores are mostly consistent with the BSF scores. Both the BSF and F1 scores show that there is a minor deviation in the scores obtained from structure learning, depending on the ordering of interventional data sets, and the standard deviation decreases with the number and size of the interventional data sets.

*5.2 Assessing the impact of Factors 1, 2, and 3, described in subsection 3.2*

We assess the impact of the three factors described in subsection 3.2 by modifying Equation (6) to consider one, or combinations of two, factors at a time. As shown in Table 7, mFGS-BS-1 refers to considering Factor 1 only, mFGS-BS-23 considers Factors 2 and 3, etc. The impact is measured in terms of graphical accuracy, based on the metrics Precision, Recall, F1 and BSF shown in Table 7. The experiments are based on the Alarm network and assume 5% latent variables (one latent variable in this case), and sample sizes 1k and 10k.

The results in Table 7 depict the average learning performance over 10 experiments, from considering just one interventional data set to considering 10 interventional data sets. We repeat these experiments five times, and each time we randomly choose a new variable to be targeted for intervention. Considering one factor alone, the results clearly show considerable drop in performance across almost all cases. Combinations of two factors increase performance, particularly when Factor 3 is included in the combination. Although Factor 1 appears to be less important than Factors 2 and 3, considering all three factors (i.e., the default mFGS-BS) does lead to a slightly better overall performance across all combinations.

| Metric | n | mFGS-BS | mFGS-BS-1 | mFGS-BS-2 | mFGS-BS-3 | mFGS-BS-12 | mFGS-BS-13 | mFGS-BS-23 |
|---|---|---|---|---|---|---|---|---|
| Precision | 1k | 0.79 | 0.45 | **0.76** | 0.58 | 0.45 | **0.82** | 0.78 |
| Recall | | 0.74 | **0.71** | 0.56 | 0.36 | 0.71 | **0.73** | 0.58 |
| F1 | | 0.77 | 0.55 | **0.64** | 0.44 | 0.55 | **0.77** | 0.66 |
| BSF | | 0.74 | **0.65** | 0.56 | 0.36 | 0.65 | **0.73** | 0.58 |
| Precision | 10k | 0.79 | 0.64 | **0.76** | 0.63 | 0.63 | **0.78** | 0.77 |
| Recall | | 0.75 | **0.74** | 0.69 | 0.55 | **0.74** | 0.74 | 0.71 |
| F1 | | 0.77 | 0.68 | **0.72** | 0.59 | 0.68 | **0.76** | 0.74 |
| BSF | | 0.75 | **0.72** | 0.69 | 0.55 | 0.71 | **0.74** | 0.70 |

**Table 7.** The impact of Factors 1, 2 and 3 (refer to subsection 3.2) on graphical performance, where mFGS-BS considers all of the three factors (default version), mFGS-BS-1 considers Factor 1 only, mFGS-BS-12 considers Factors 1 and 2 only, etc. The results represent average performance over multiple experiments with synthetic Alarm network data, as described in section 5.2.

*5.3 Results based on one variable targeted for intervention per interventional data set*

In this subsection, we assume that each interventional data contains a single variable that is randomly targeted for intervention. Because RFCI-BSC failed to generate a PAG for almost all cases in which the sample size is 10k, we restrict its comparisons to experiments where the sample size is up to 1k. Figure 6 shows the results obtained by applying the algorithms to the Asia network over two sample sizes. The x-axis represents the total number of interventional data sets considered for learning, and the y-axis represents the specified scoring metric, runtime, or the number of edges learnt. Each data point in these graphs represents the average result across five iterations. Each iteration involves new data sets and new variables targeted for intervention. The results show that mFGS-BS outperforms GFCI and RFCI-BSC, and to a lesser degree COmbINE which demonstrates erratic performance, across all four metrics and two sample sizes. Importantly, the results show that both mFGS-BS and COmbINE continue to improve with the number of interventional data sets. Conversely, the graphical accuracy of GFCI and RFCI-BSC decreases with the number of interventional data sets, and this is expected since these two algorithms used pooled data, where the post-interventional and pre-interventional distributions may conflict. Lastly, COmbINE is found to be considerably faster than both mFGS-BS and GFCI at 10k sample size.

Figure 7 repeats the results for the Sports network, which is also a small network. However, compared to Asia, the Sports network contains a considerably higher number of free parameters. Overall, the results show that the algorithms deliver a rather similar performance when the number of data sets is low, with the gap in performance increasing as the number of data sets increases. The accuracy of mFGS-BS increases faster with the number of data sets, and this eventually makes the gap in performance important at higher number of data sets. Interestingly, while COmbINE is the fastest algorithm on Asia, it is the slowest on Sports. A possible explanation is the number of free parameters, which is 1,049 in Sports compared to just 18 in Asia, despite the two networks having just one variable difference. This suggests that COmbINE might not scale well with dense networks, or with networks that contain multinomial rather than Boolean variables, whereas RFCI-BSC fails to return an output and instead returns an out-of-memory error. Lastly, GFCI produces a high number of learnt edges, and this number continues to increase with the number of data sets and greatly surpasses the number of true edges.

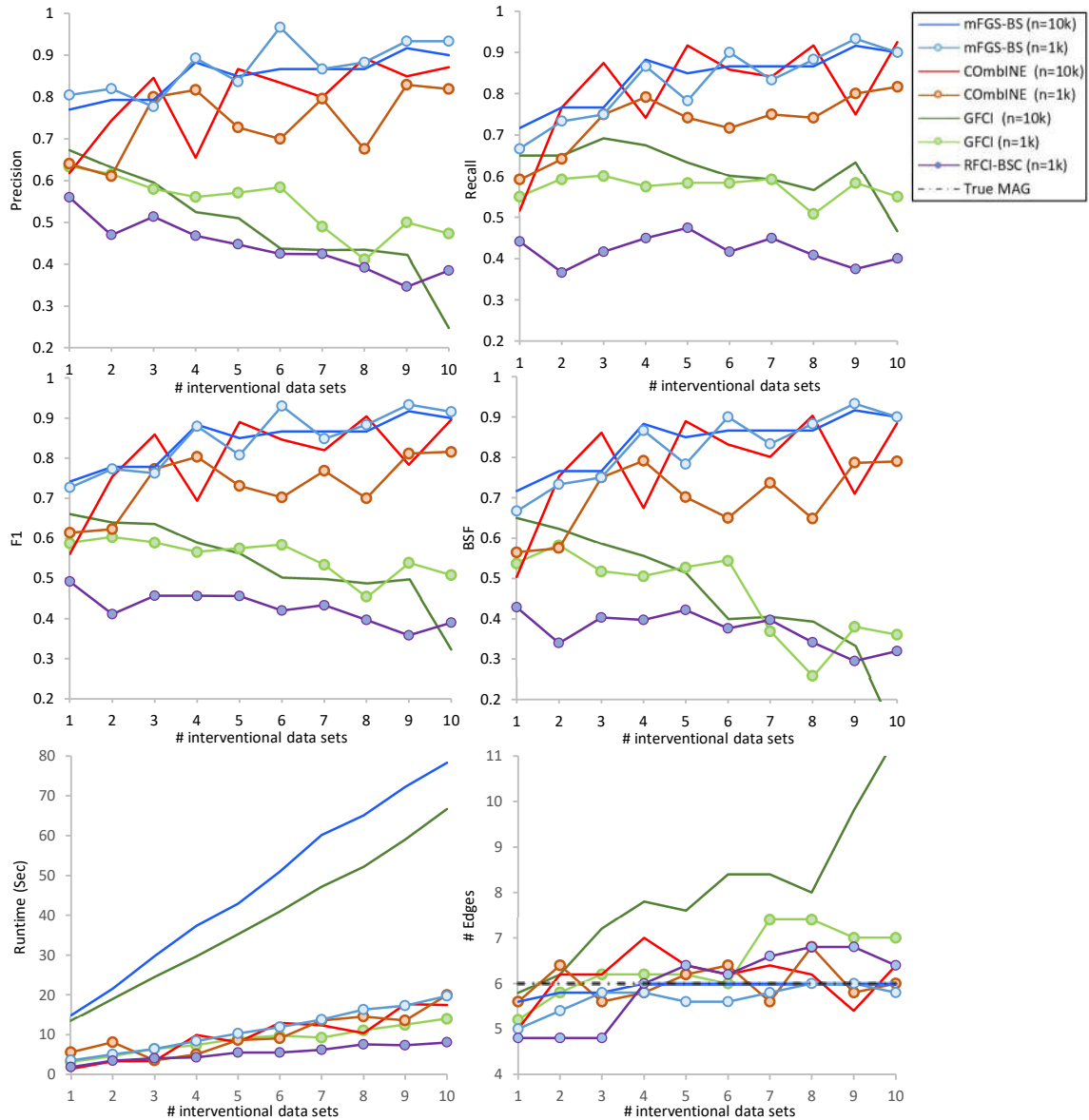

**Figure 6.** Average performance of the algorithms when applied to synthetic data generated from the Asia network, assuming one intervened variable and 10% latent variables per data set, over two sample sizes.

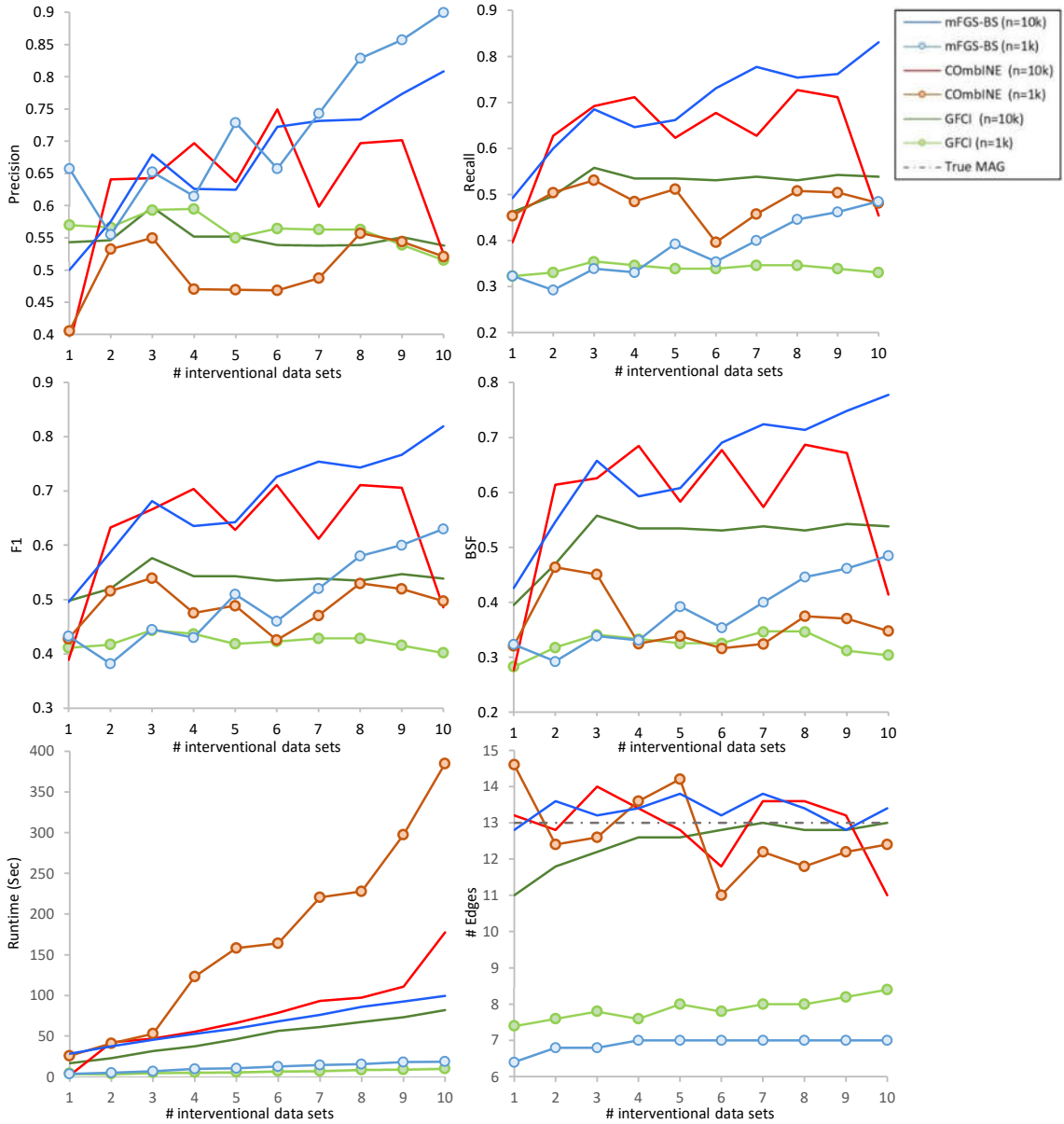

**Figure 7**. Average performance of the algorithms when applied to synthetic data generated from the Sports network, assuming one intervened variable and 10% latent variables per data set, over two sample sizes.

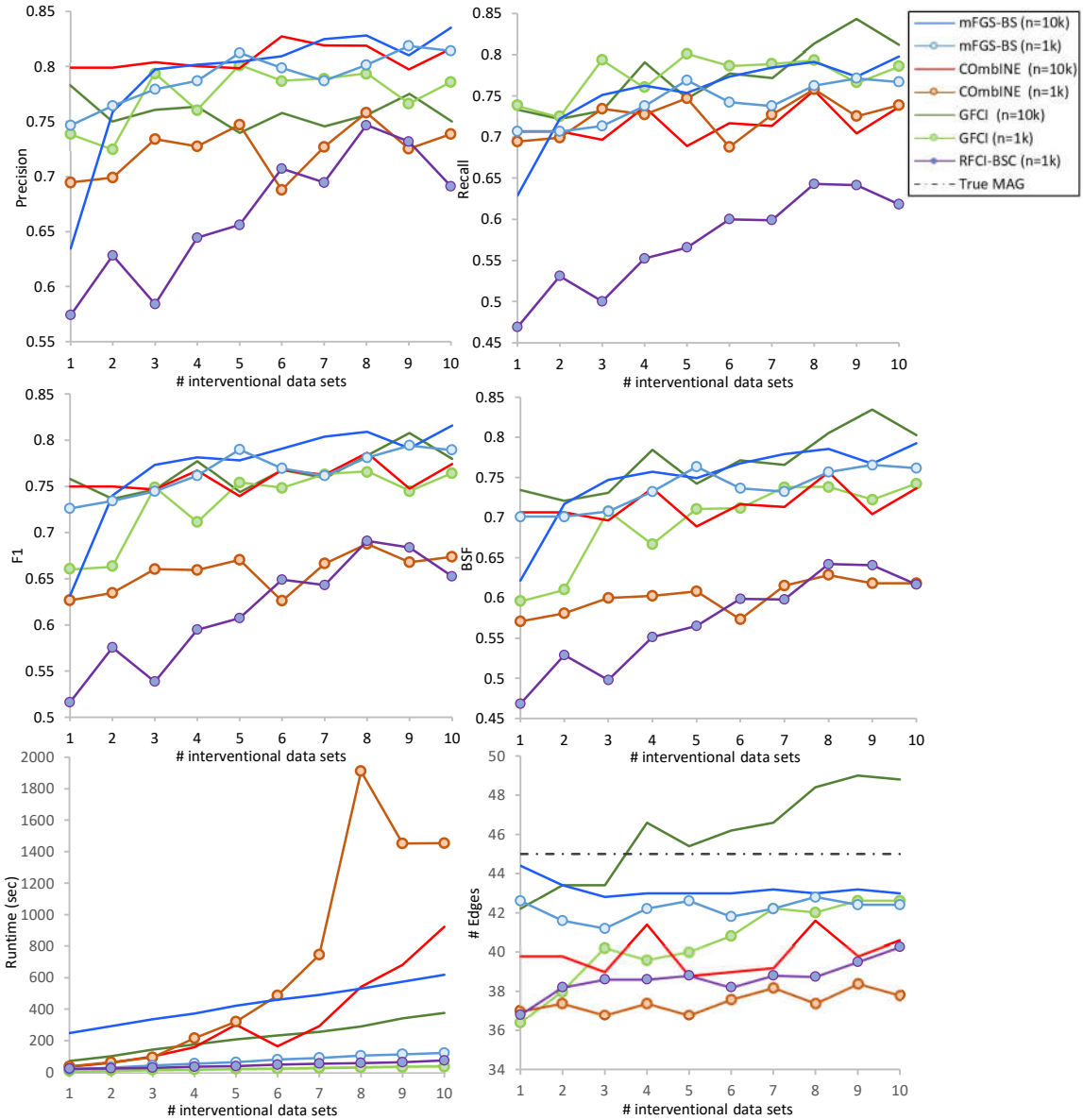

**Figure 8.** Average performance of the algorithms when applied to synthetic data generated from the Alarm network, assuming one intervened variable and 5% latent variables per data set, over two sample sizes.

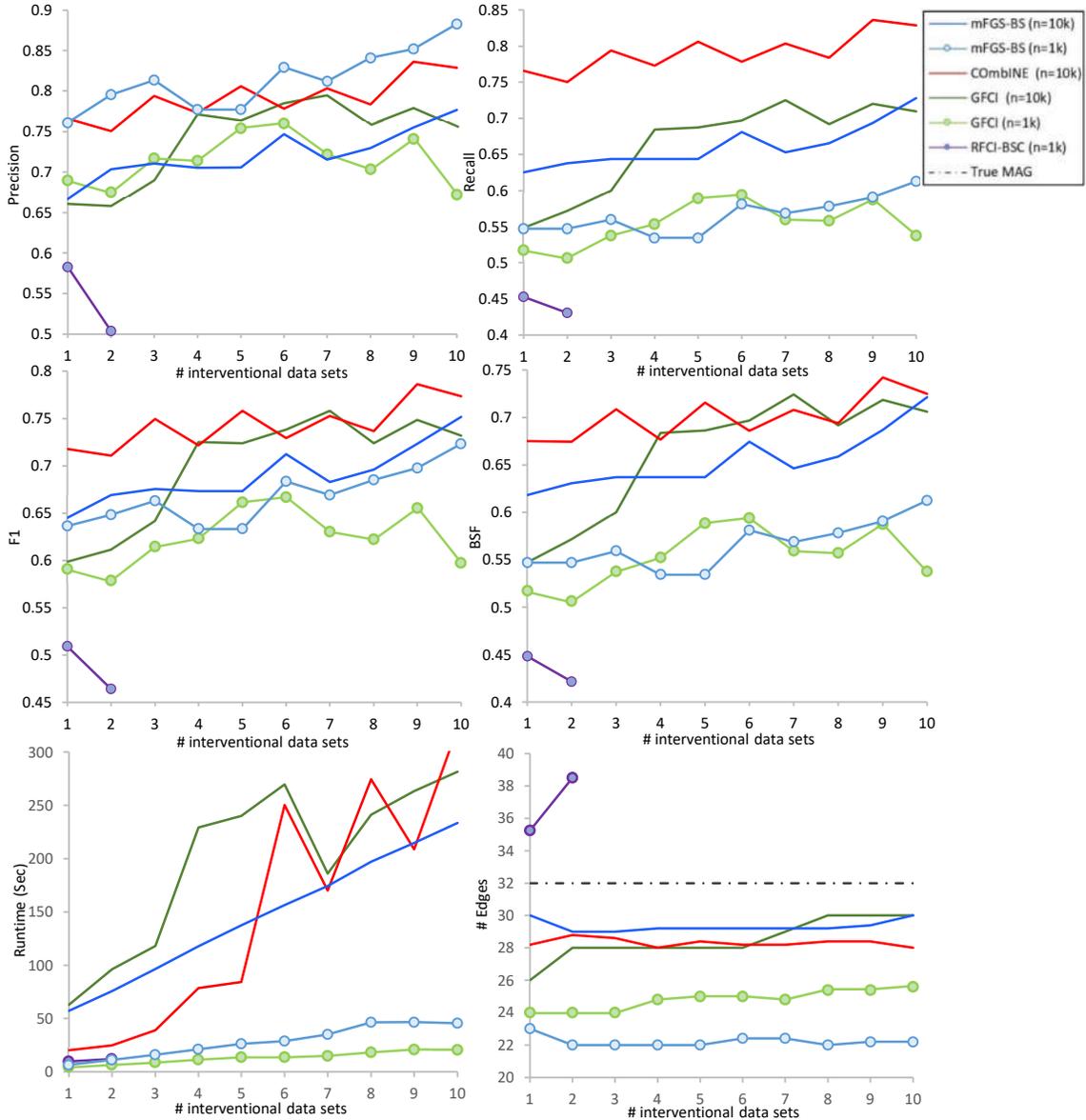

**Figure 9.** Average performance of the algorithms when applied to synthetic data generated from the Property network, assuming one intervened variable and 5% latent variables per data set, over two sample sizes.

Table 8 summarises the average results across all experiments in which a single variable is targeted for intervention. The results show that mFGS-BS performed best in the small and medium networks and across all four scoring metrics, followed by COmbINE, then GFCI and finally RFCI-BSC. In terms of runtime, however, GFCI is found to be the fastest algorithm in most experiments, followed by mFGS-BS, then COmbINE, and finally RFCI-BSC which could not process any of the larger networks within the runtime limit.

Figures 8 and 9 repeat the results for the medium networks Alarm and Property respectively. While there are some variations in the results, the overall conclusions that can be derived from these results are consistent with those derived from the smaller networks of Asia and Sports. A notable exception is that COmbINE performs better than mFGS-BS, in terms of BSF and recall, in Property. However, this result is restricted to the sample size of 10k, and this is because COmbINE fails to generate a result within the four-hour runtime limit for sample size 1k and RFCI-BSC fails to return a result when the experiments rely on more than two interventional data sets. Because COmbINE does not return a result for any of these larger networks within the four-hour time limit, we shown these results in the Appendix (see Figures 12 and 13). Overall, the larger networks show that GFCI outperforms mFGS-BS slightly in ForMed, perhaps because any differences between the observational and interventional data with just one intervened node is relatively minor in this larger network. mFGS-BS outperforms GFCI considerably in Pathfinder in terms of graphical accuracy. Pathfinder is the network with the highest number of free parameters considered in this study, and this complexity might explain why all algorithms perform relatively poorly on Pathfinder compared to the other networks.

| Algorithm | n | Asia | Sports | Property | Alarm | ForMed | Pathfinder | Asia | Sports | Property | Alarm | ForMed | Pathfinder |
|---|---|---|---|---|---|---|---|---|---|---|---|---|---|
| | | | | | | **Precision** | | | | | | **Recall** | |
| mFGS-BS | 1k | **0.87** | **0.72** | **0.81** | 0.79 | **0.90** | 0.29 | 0.83 | 0.38 | **0.57** | **0.74** | 0.46 | **0.16** |
| | 10k | **0.85** | **0.68** | 0.66 | 0.79 | **0.79** | 0.50 | **0.84** | **0.69** | 0.64 | 0.75 | 0.67 | **0.32** |
| COmbINE | 1k | 0.74 | 0.50 | T | 0.72 | T | T | 0.73 | 0.48 | T | 0.60 | T | T |
| | 10k | 0.80 | 0.63 | **0.79** | **0.81** | T | T | 0.81 | 0.62 | **0.70** | 0.72 | T | T |
| GFCI | 1k | 0.54 | 0.56 | 0.71 | 0.77 | 0.75 | 0.16 | 0.57 | 0.34 | 0.55 | 0.70 | **0.56** | 0.11 |
| | 10k | 0.49 | 0.55 | 0.74 | 0.76 | 0.77 | 0.12 | 0.62 | 0.53 | 0.66 | **0.77** | **0.71** | 0.11 |
| RFCI-BSC | 1k | 0.44 | M | 0.54 | 0.67 | T | T | 0.42 | M | 0.44 | 0.57 | T | T |
| | | | | | | **F1** | | | | | | **BSF** | |
| mFGS-BS | 1k | **0.85** | **0.50** | **0.67** | **0.77** | 0.61 | **0.20** | **0.83** | **0.38** | **0.57** | **0.74** | 0.46 | **0.14** |
| | 10k | **0.84** | **0.68** | 0.65 | 0.77 | 0.72 | **0.39** | **0.84** | **0.65** | 0.63 | 0.75 | 0.66 | **0.31** |
| COmbINE | 1k | 0.73 | 0.49 | T | 0.66 | T | T | 0.70 | 0.36 | T | 0.60 | T | T |
| | 10k | 0.80 | 0.62 | **0.74** | 0.76 | T | T | 0.78 | 0.58 | **0.70** | 0.72 | T | T |
| GFCI | 1k | 0.55 | 0.42 | 0.62 | 0.73 | **0.64** | 0.13 | 0.46 | 0.32 | 0.55 | 0.69 | **0.56** | 0.09 |
| | 10k | 0.54 | 0.54 | 0.70 | **0.77** | **0.73** | 0.12 | 0.46 | 0.52 | 0.66 | **0.77** | **0.70** | 0.08 |
| RFCI-BSC | 1k | 0.43 | M | 0.49 | 0.62 | T | T | 0.37 | M | 0.43 | 0.57 | T | T |
| | | | | | | **Learnt Edges** | | | | | | **Runtime** | |
| mFGS-BS | 1k | 5.68 | 6.90 | 22.22 | 42.18 | 72.32 | 124.52 | 11.26 | 11.35 | 28.43 | 72.21 | 333.68 | 1026.19 |
| | 10k | 5.92 | 13.34 | 31.00 | 43.20 | 117.86 | 148.28 | 47.31 | 64.44 | 213.08 | 434.71 | 1556.64 | 2756.12 |
| COmbINE | 1k | 6.02 | 12.70 | T | 37.48 | T | T | 10.14 | 169.59 | T | 677.86 | T | T |
| | 10k | 6.14 | 12.94 | 28.32 | 39.90 | T | T | **9.65** | 76.89 | **147.34** | 325.85 | T | T |
| GFCI | 1k | 6.44 | 7.88 | 24.80 | 40.44 | 105.72 | 161.52 | 8.68 | **6.66** | **14.19** | **21.90** | **39.12** | **63.03** |
| | 10k | 8.06 | 12.46 | 28.58 | 46.00 | 129.90 | 210.04 | 38.76 | 49.39 | 162.28 | 219.94 | 1014.59 | 548.98 |
| RFCI-BSC | 1k | 5.96 | M | 36.88 | 38.65 | T | T | **5.37** | M | M | 44.59 | T | T |

**Table 8.** Average performance across all experiments in which a single variable is targeted for intervention per data set, where M indicates out-of-memory error, and T indicates failure to complete learning within the four-hour runtime limit. The best performance values are shown in bold.

### 5.4 Results based on five variables targeted for intervention per interventional data set

This subsection focuses on the results when the number of intervened variables is increased from one (the results in subsection 5.3) to five, for each interventional data. Because the Asia and Sports networks contain less than 10 variables, we do not consider them here since it would be unrealistic to assume that half of the network variables are targeted for intervention. Instead, we consider the networks of Property, Alarm, ForMed and Pathfinder where the number of variables ranges from 27 to 109.

Figure 10 presents the results based on the Property network and shows that both mFGS-BS and COmbINE improve their performance relative to the corresponding results in Figure 9 which consider only one intervened variable. Table 9, which summarises the average results obtained when considering five intervened variables, shows that mFGS-BS performs best across all metrics at 1k sample size, whereas COmbINE performs best across all metrics at 10k sample size for the Property network. However, as shown in Figure 10, the runtime of COmbINE increases much faster with the number of data sets, and fails to generate any results within the four-hour runtime limit when the number of data sets is three or more. RFCI-BSC, on the other hand, returned an out-of-memory error when applied to these data sets. Therefore, the average results reported in Table 9 may underestimate the performance of COmbINE and RFCI-BSC for sample size 1k, since the average is derived solely by focusing on a lower number of data sets on which the performance tends to be worse.

Figure 11 repeats the results for the Alarm network. As before, COmbINE failed to produce a result for all experiments within the four-hour time limit. However, the results of COmbINE this time extend up to six interventional data sets and enable us to draw reasonably confident conclusions. mFGS-BS performs best overall and across almost all the different number of data sets and sample sizes. Both mFGS-BS and COmbINE perform better compared to the case of a single intervened variable, and continue to improve with the number of data sets, whereas GFCI and RFCI-BSC do not.

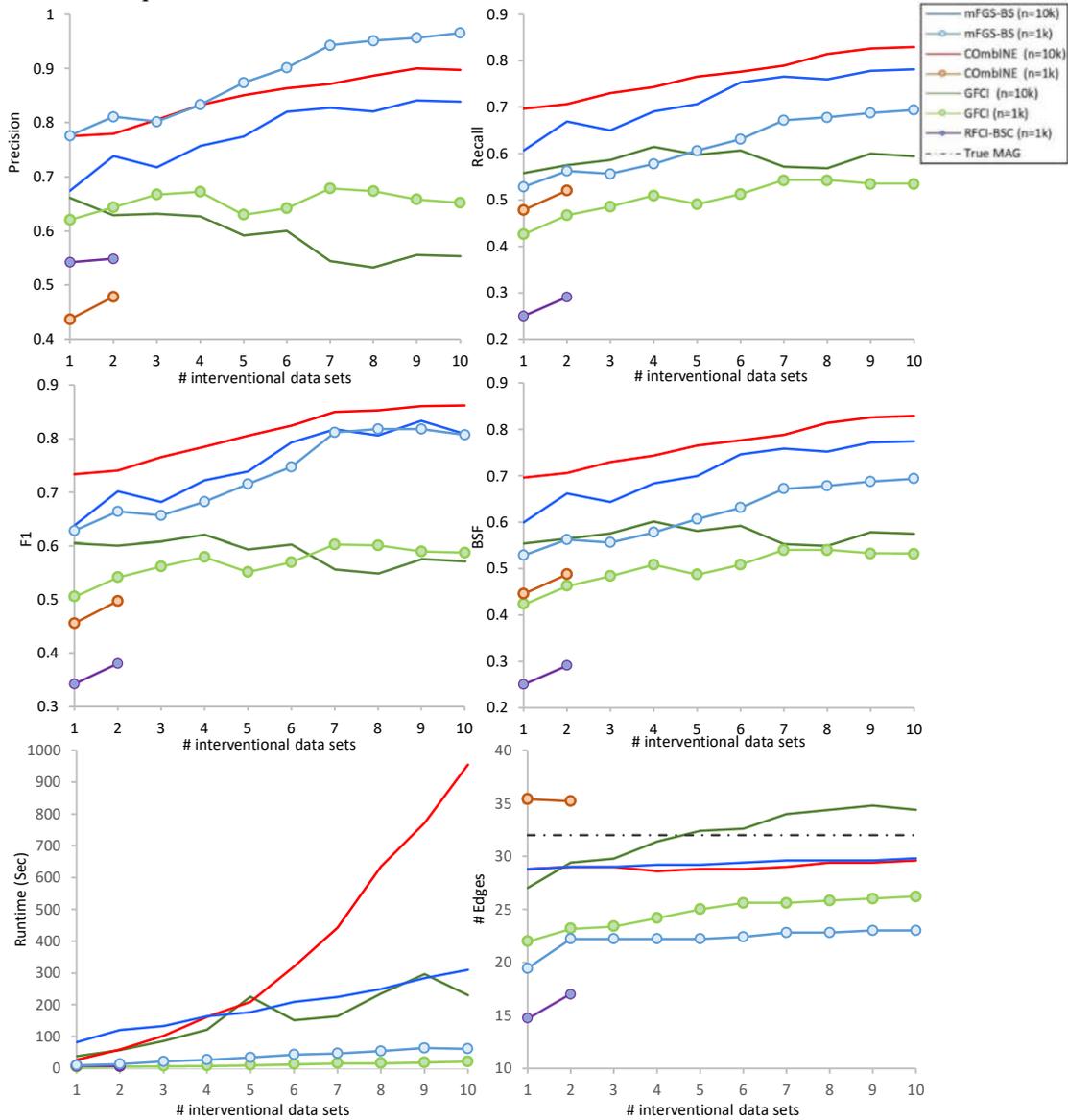

**Figure 10**. Average performance of the algorithms when applied to synthetic data generated from the Property network, assuming five intervened variables and 5% latent variables per data set, over two sample sizes. The runtime of COmbINE at 1k sample size is not shown in the charts, because its runtime is much higher.

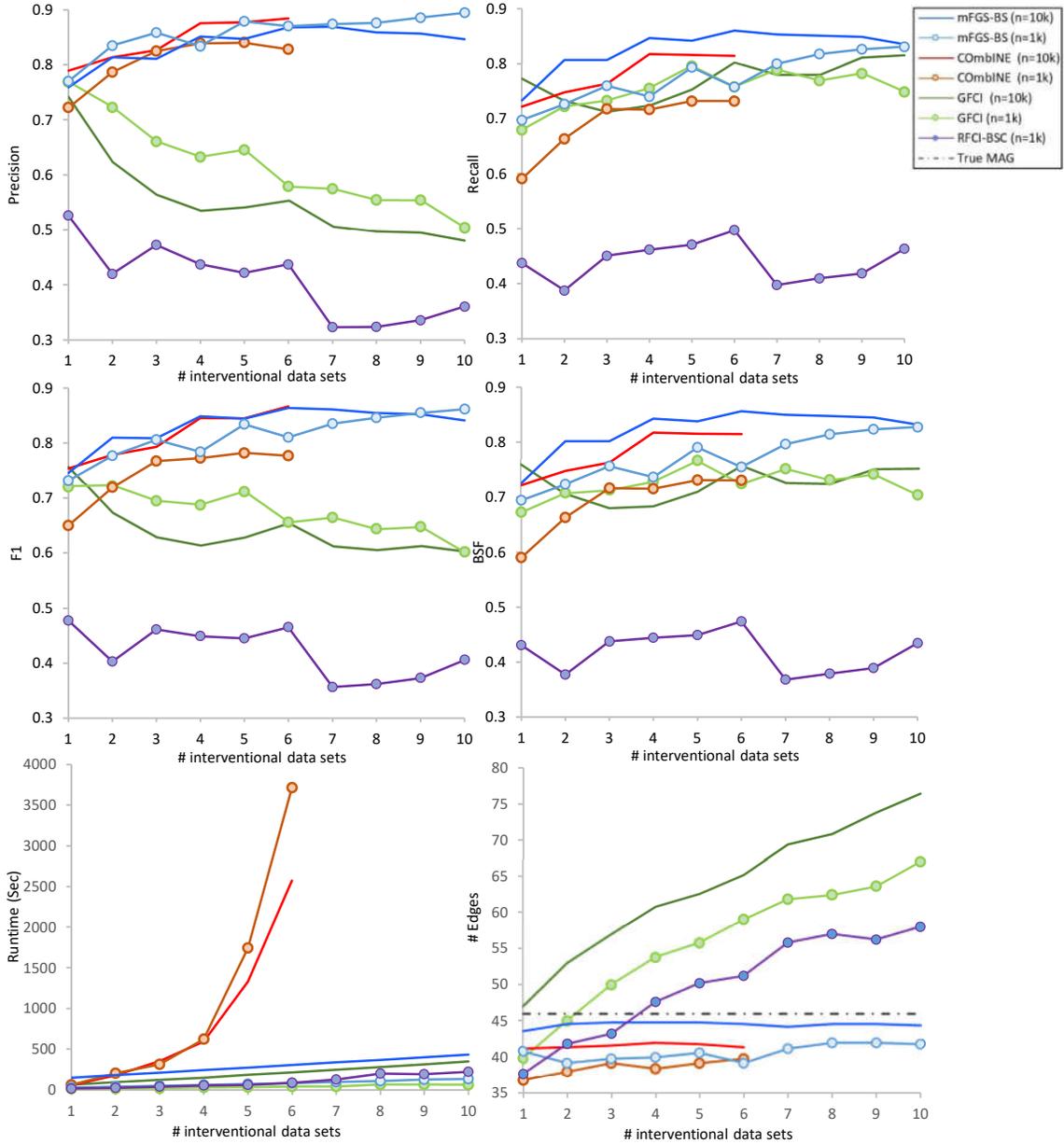

**Figure 11**. Average performance of the algorithms when applied to synthetic data generated from the Alarm network, assuming five intervened variables and 5% latent variables per data set, over two sample sizes.

For the large and very large networks, COmbINE and RFCI-BSC fail to produce any results. On the other hand, both mFGS-BS and GFCI are able to generate results for all experiments across both sample sizes. The experimental results obtained from ForMed and Pathfinder case studies can be found in the Appendix (Figures 14 and 15). Note that, in the case of these larger networks, five intervened variables represent a relatively low number. Still, as shown in Table 9, mFGS-BS performs considerably better than GFCI and RFCI-BSC across almost all experiments. The only case in which GFCI performs slightly better than mFGS-BS is for ForMed at 1k sample size, where GFCI averages scores of 0.58 and 0.59 for BSF and Recall respectively, whereas mFGS-BS averages scores of 0.57 for both metrics. On the other hand, the cases in which mFGS-BS outperforms GFCI involve much higher discrepancies in scores. For example, the most extreme case involves the Pathfinder case study where mFGS-BS averages a Precision score of 0.52 at 10k sample size, whereas GFCI averages a score of just 0.13.

| Algorithms | n | Property | Alarm | ForMed | Pathfinder | Property | Alarm | ForMed | Pathfinder |
|---|---|---|---|---|---|---|---|---|---|
| | | \multicolumn{4}{c}{Precision} | \multicolumn{4}{c}{Recall} |
| mFGS-BS | 1k | **0.88** | **0.86** | **0.83** | **0.31** | 0.62 | **0.78** | 0.57 | **0.17** |
| | 10k | 0.78 | 0.82 | **0.80** | **0.52** | 0.72 | **0.79** | **0.69** | **0.35** |
| COmbINE | 1k | 0.46 | 0.81 | T | T | 0.50 | 0.69 | T | T |
| | 10k | **0.85** | **0.84** | T | T | **0.77** | 0.78 | T | T |
| GFCI | 1k | 0.65 | 0.62 | 0.74 | 0.17 | 0.50 | 0.75 | **0.59** | 0.13 |
| | 10k | 0.59 | 0.55 | 0.69 | 0.14 | 0.59 | 0.77 | 0.67 | 0.13 |
| RFCI-BSC | 1k | 0.55 | 0.41 | T | T | 0.27 | 0.44 | T | T |
| | | \multicolumn{4}{c}{F1} | \multicolumn{4}{c}{BSF} |
| mFGS-BS | 1k | **0.74** | **0.81** | 0.68 | **0.22** | 0.62 | **0.77** | 0.57 | **0.15** |
| | 10k | 0.75 | **0.81** | **0.74** | **0.42** | 0.71 | **0.79** | **0.69** | **0.34** |
| COmbINE | 1k | 0.48 | 0.74 | T | T | 0.47 | 0.69 | T | T |
| | 10k | **0.81** | **0.81** | T | T | **0.77** | 0.78 | T | T |
| GFCI | 1k | 0.57 | 0.68 | 0.65 | 0.14 | 0.50 | 0.72 | **0.58** | 0.10 |
| | 10k | 0.59 | 0.64 | 0.68 | 0.13 | 0.57 | 0.73 | 0.66 | 0.10 |
| RFCI-BSC | 1k | 0.36 | 0.42 | T | T | 0.27 | 0.42 | T | T |
| | | \multicolumn{4}{c}{Learnt Edges} | \multicolumn{4}{c}{Runtime} |
| mFGS-BS | 1k | 22.22 | 40.66 | 95.54 | 126.96 | 37.69 | 77.98 | 413.96 | 762.02 |
| | 10k | 29.32 | 43.44 | 121.42 | 152.32 | 195.36 | 417.33 | 2411.11 | 2737.49 |
| COmbINE | 1k | 35.30 | 38.57 | T | T | 5420.72 | 1108.76 | T | T |
| | 10k | 29.04 | 41.57 | T | T | 368.40 | 845.34 | T | T |
| GFCI | 1k | 24.70 | 55.82 | 111.40 | 171.36 | **12.91** | **34.80** | **55.19** | **66.45** |
| | 10k | 32.02 | 63.60 | 136.80 | 220.00 | **160.72** | **199.41** | **478.82** | **546.57** |
| RFCI-BSC | 1k | 15.88 | 49.86 | T | T | M | 99.8 | T | T |

**Table 9.** Average performance across all experiments in which five variables are targeted for intervention per data set, where M indicates out-of-memory error, and T indicates failure to complete learning within the four-hour runtime limit. The best performance values are shown in bold.

The main conclusions from the results are:

- mFGS-BS is found to be sensitive to the ordering of interventional data sets. However, the sensitivity is relatively small in terms of graphical accuracy, and decreases with the number and the size of interventional data sets.

- Employing all three factors to determine edge direction produces the most accurate graphs (refer to subsection 3.2.2). Factor 1, which determines the probability of directed edges given the output of FGS, and Factor 2 which determines the probability of directed edges based on the ratio of Sepsets determining v-structure, are found to have a stronger impact (in terms of increasing the F1 and BSF scores) than Factor 3 which relies on changes in objective score between observational and interventional data.

- mFGS-BS is found to be more accurate than the other algorithms when we simulate just one intervened variable. Specifically, mFGS-BS generates the highest F1 and BSF scores for the Asia, Sports and Alarm networks in most of the experiments (refer to Table 8). COmbINE and RFCI-BSC often fail to generate a result within the four-hour runtime limit when applied to the larger networks. The average BSF and F1 scores of mFGS-BS are approximately 45% and 38% higher compared to GFCI across all networks, while the average BSF and F1 scores of COmbINE are 16% and 15% higher compared to GFCI over all experiments in which COmbINE generates a result.

- The performance of both mFGS-BS and COmbINE continues to improve with the number of interventional data sets, while the performance of GFCI and RFCI-BSC does not. This highlights the advantage of algorithms that consider additional data sets independently. Moreover, the number of edges learnt by mFGS-BS tends to be lower compared to the number of edges present in the true MAGs, for the medium, large and very large networks. Note that while GFCI generates more edges when the number of interventional data sets increase, its overall performance in terms of BSF and F1 scores does not increase.

- The overall performance of mFGS-BS and COmbINE continues to improve with the number of variables targeted for intervention as expected, since the higher number of interventions can be viewed as providing additional causal information to the model. The average BSF scores increase by approximately 9% and 11% when considering five, instead of one, intervened variables per interventional data for the mFGS-BS and COmbINE algorithms respectively.

- The runtime of mFGS-BS, relative to the other three algorithms, appears to be worst in small and medium networks. However, the runtime of mFGS-BS, RFCI-BSC and GFCI scale linearly with the number of interventional data sets. In contrast, the empirical results suggest that COmbINE does not scale well with additional data sets. One explanation might be because COmbINE uses the MINISAT application to solve SAT instances encoded from results of CI tests, and the time to solve these SAT instances increases exponentially with the number of variables. A rather unexpected finding is that the computational time of COmbINE is higher when the sample size is 1k compared to 10k. This might be because the results of CI tests learnt from low sample sizes contain more conflicts compared to those obtained when the sample size of the input data is higher. Lastly, GFCI is found to be the fastest algorithm in almost all of the experiments, as expected, since it does not consider each input data set independently.

## 6. Conclusion

This paper describes the mFGS-BS hybrid algorithm which produces a PAG by learning the probabilities of each directed edge from one observational data set and one or more interventional data sets in a causally insufficient setting. The posterior probabilities learnt from one data set are considered as candidate objective priors for learning from the next data set. Three other mechanisms contribute to the objective priors used with each data set: colliders identified from the observational data; the CPDAGs produced by running the FGS algorithm on each data set; and a score-based approach relating to intervention targets. Pairs of nodes which have a directed edge in each direction with a probability above a given threshold are treated as having a bidirected edge between them, so that the algorithm produces a PAG.

The results of mFGS-BS were compared to those obtained by COmbINE, which also enables learning from multiple observational and interventional data sets. We have also compared the results against the RFCI-BSC and GFCI algorithms with pooled data, which serves as the baseline performance not accounting for variables targeted for intervention. The empirical evaluation was based on six case studies of different complexity, with varying numbers of intervened variables, interventional data sets, and sample sizes. Overall, the results show that mFGS-BS considerably outperforms the baseline algorithms in terms of graphical accuracy, and also outperforms COmbINE in most of the experiments. RFCI-BSC and GFCI consider a single data set of pooled data rather than each input data set independently. GFCI was the faster algorithm because it performs fewer CI tests by design, whereas RFCI-BSC tends to fail to produce a result when applied to larger networks and sample sizes. Lastly, mFGS-BS offers considerable improvements in learning efficiency compared to COmbINE, which failed to produce any results, within the four-hour runtime limit, for the larger networks.

A limitation of mFGS-BS is that it is sensitive to the ordering of the data sets and assumes equal sample size across all input data sets. This is, of course, an unrealistic assumption in practice. Future revisions of mFGS-BS will adjust the algorithm such that the local BDeu scores can be normalised to enable learning from multiple data sets with varying sample sizes. Other future research directions could focus on enabling learning from interventional data sets that contain imperfect and uncertain interventions (refer to subsection 2.2), in addition to perfect interventions.

**Acknowledgements**

This research was supported by the ERSRC Fellowship project EP/S001646/1 on *Bayesian Artificial Intelligence for Decision Making under Uncertainty* (Constantinou, 2018), and by the Royal Thai Government Scholarship offered by Thailand's Office of Civil Service Commission (OCSC).

# Appendix: Supplementary results

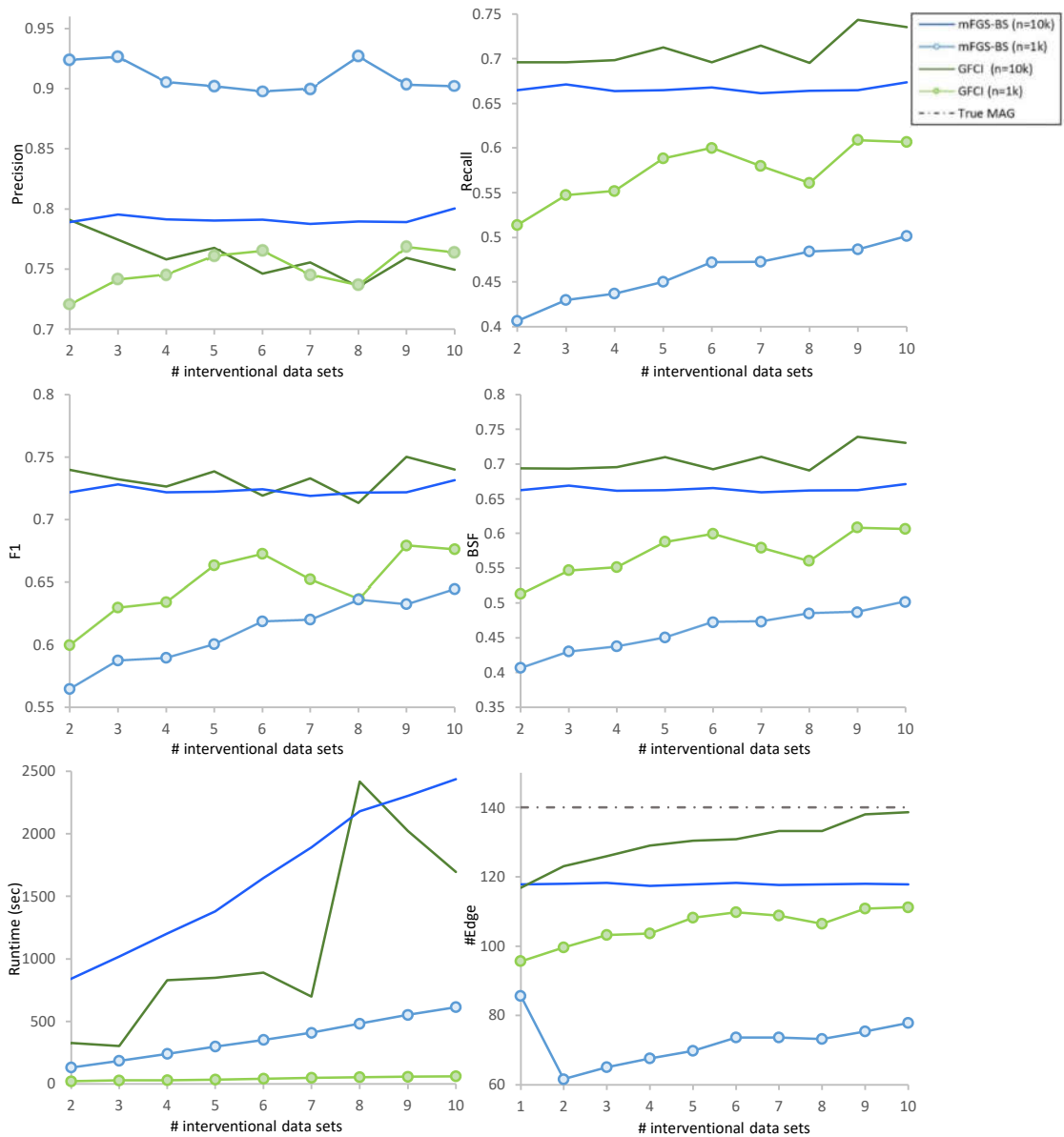

**Figure 12**. Average performance of the algorithms when applied to synthetic data generated from the Formed network, assuming one intervened variable and 5% latent variables per data set, over two sample sizes.

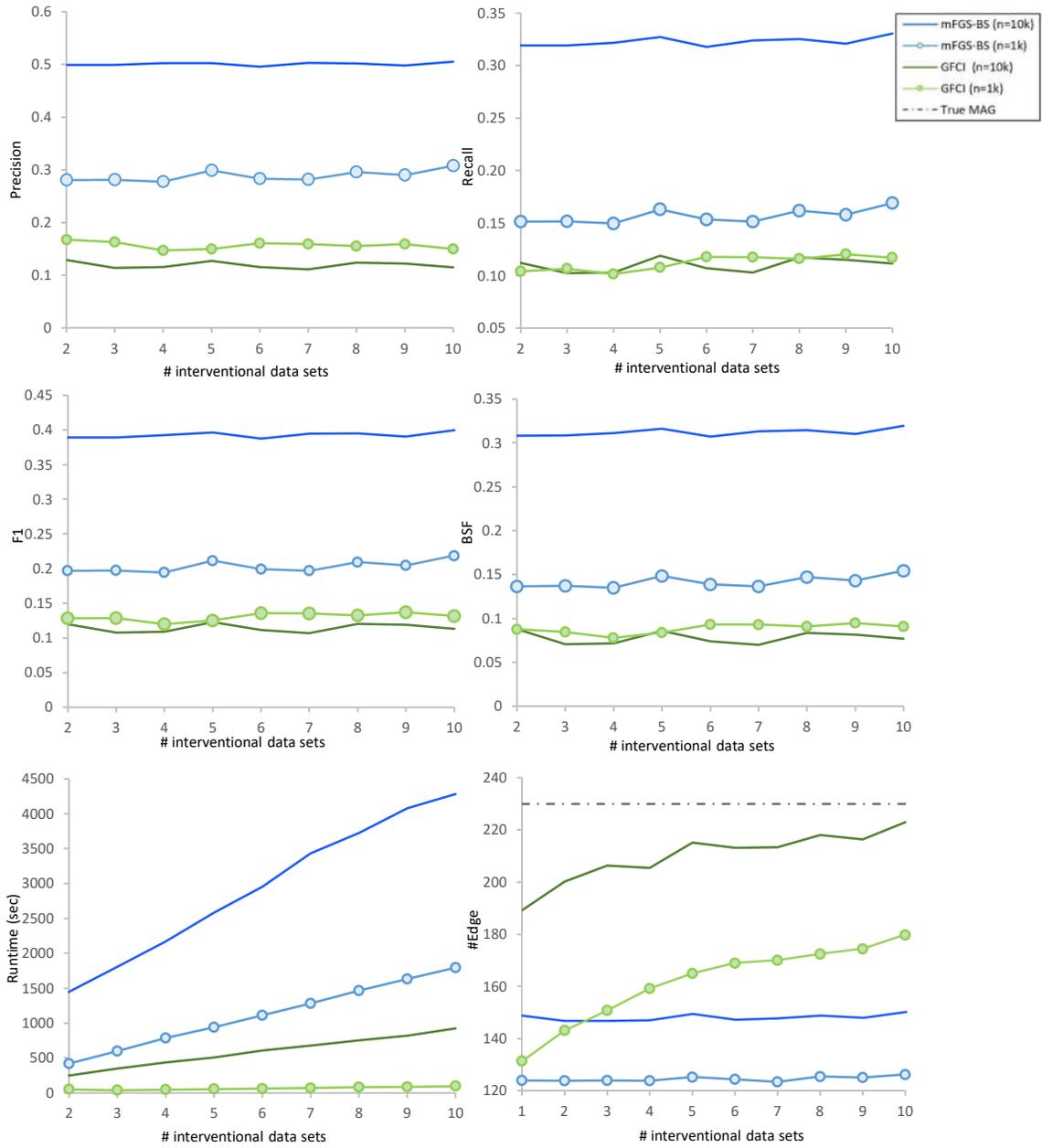

**Figure 13**. Average performance of the algorithms when applied to synthetic data generated from the Pathfinder network, assuming one intervened variable and 5% latent variables per data set, over two sample sizes.

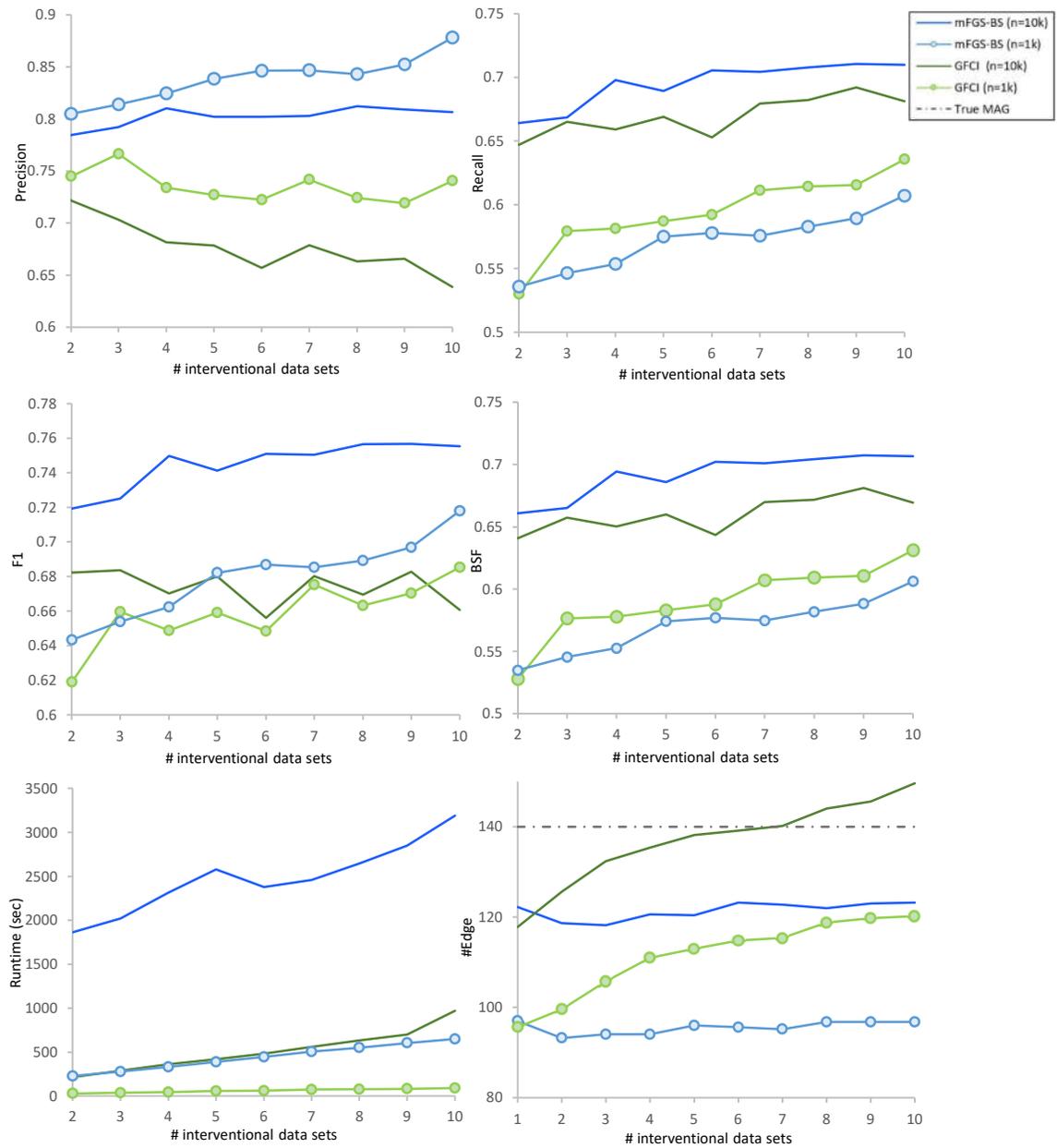

**Figure 14**. Average performance of the algorithms when applied to synthetic data generated from the Formed network, assuming five intervened variables and 5% latent variables per data set, over two sample sizes.

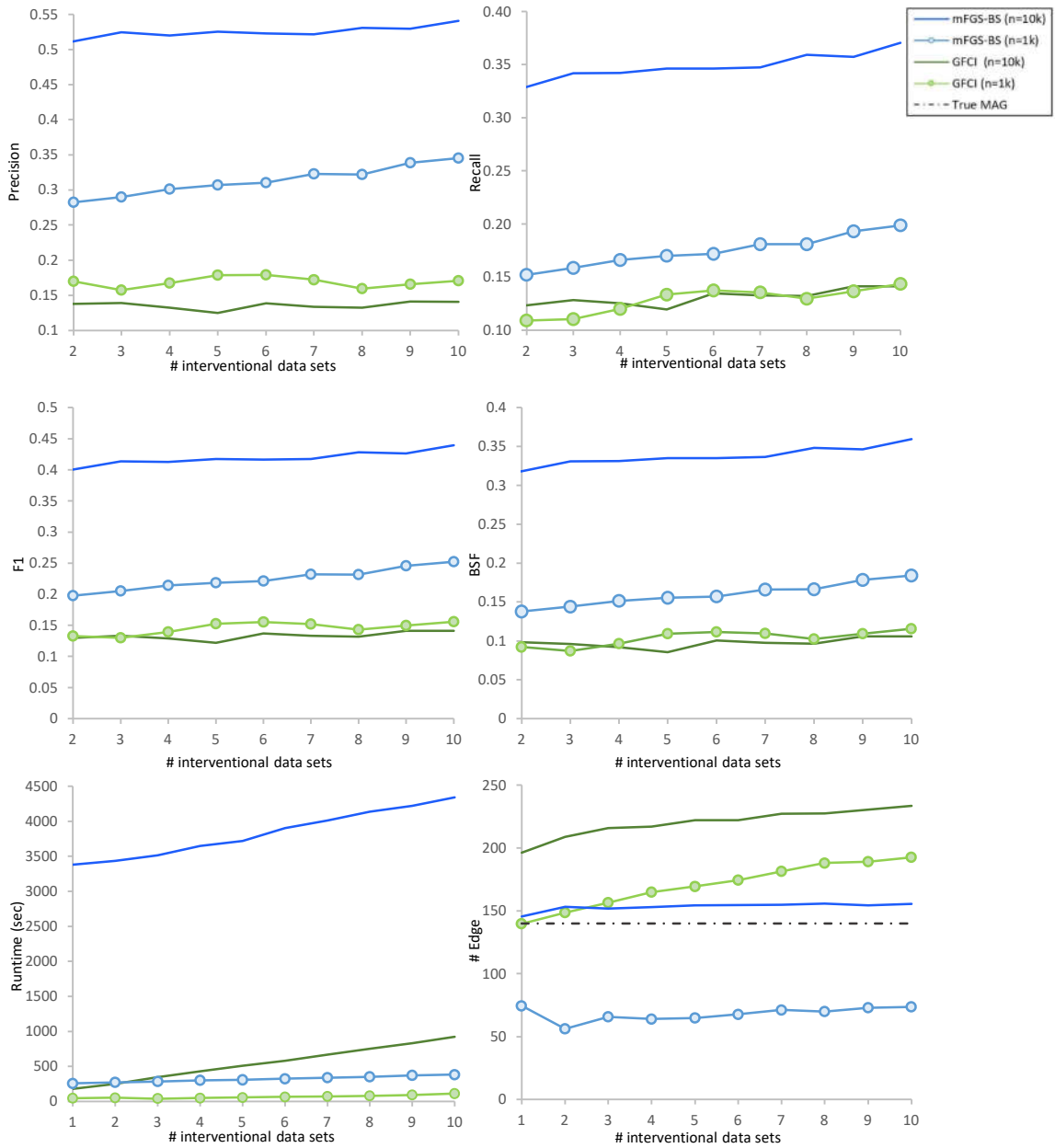

**Figure 15**. Average performance of the algorithms when applied to synthetic data generated from the Pathfinder network, assuming five intervened variables and 5% latent variables per data set, over two sample sizes.